\documentclass[10pt,journal,compsoc]{IEEEtran}

\ifCLASSOPTIONcompsoc
  \usepackage[nocompress]{cite}
\else
  \usepackage{cite}
\fi
\ifCLASSINFOpdf
\else
\fi

\usepackage{amsmath}
\usepackage{amssymb}
\usepackage[numbers]{natbib}
\usepackage{xcolor}
\usepackage{comment}
\usepackage{amsmath,amssymb} 
\usepackage{color}
\usepackage{hyperref}
\usepackage{enumitem}
\usepackage{adjustbox}
\usepackage{subcaption}
\usepackage{amsmath}
\usepackage{amssymb}
\usepackage{booktabs}
\usepackage{caption,fixltx2e}
\usepackage[flushleft]{threeparttable}
\usepackage{pifont}
\usepackage{algorithm}
\usepackage{tabu}
\usepackage{algpseudocode}
\usepackage{graphicx}
\usepackage{multirow}
\usepackage{subcaption}
\usepackage{bbm}
\newcommand{\xmark}{\ding{55}}%

\begin{document}
%
\title{Teacher-Student Architecture for Knowledge Distillation: A Survey}
%
%
%
%

\author{Chengming Hu,~\IEEEmembership{}
        Xuan Li,~\IEEEmembership{}
        Dan Liu,~\IEEEmembership{}
        Haolun Wu,~\IEEEmembership{}
        Xi Chen,~\IEEEmembership{}
        Ju Wang,~\IEEEmembership{}
        and Xue Liu~\IEEEmembership{Fellow,~IEEE}
\IEEEcompsocitemizethanks{
\IEEEcompsocthanksitem Chengming Hu, Xuan Li, Dan Liu, Haolun Wu, Xi Chen, and Xue Liu are with the School of Computer Science, McGill University, Montreal, QC, H3A 0G4, Canada \protect \\
E-mail: chengming.hu@mail.mcgill.ca; xuan.li2@mail.mcgill.ca; daniel.liu@mail.mcgill.ca; haolun.wu@mail.mcgill.ca; xi.chen11@mcgill.ca; xueliu@cs.mcgill.ca
\IEEEcompsocthanksitem Ju Wang is with the School of Information Science and Technology, Northwest University, Xi'an, China \protect\\
E-mail: wangju@nwu.edu.cn}
}

\IEEEtitleabstractindextext{
\begin{abstract}
Although Deep neural networks (DNNs) have shown a strong capacity to solve large-scale problems in many areas, such DNNs are hard to be deployed in real-world systems due to their voluminous parameters. 
To tackle this issue, Teacher-Student architectures were proposed, where simple student networks with a few parameters can achieve comparable performance to deep teacher networks with many parameters. 
Recently, Teacher-Student architectures have been effectively and widely embraced on various knowledge distillation (KD) objectives, including knowledge compression, knowledge expansion, knowledge adaptation, and knowledge enhancement. 
With the help of Teacher-Student architectures, current studies are able to achieve multiple distillation objectives through lightweight and generalized student networks.   
Different from existing KD surveys that primarily focus on knowledge compression, this survey first explores Teacher-Student architectures across multiple distillation objectives. 
This survey presents an introduction to various knowledge representations and their corresponding optimization objectives. 
Additionally, we provide a systematic overview of Teacher-Student architectures with representative learning algorithms and effective distillation schemes. 
This survey also summarizes recent applications of Teacher-Student architectures across multiple purposes, including classification, recognition, generation, ranking, and regression.
Lastly, potential research directions in KD are investigated, focusing on architecture design, knowledge quality, and theoretical studies of regression-based learning, respectively.
Through this comprehensive survey, industry practitioners and the academic community can gain valuable insights and guidelines for effectively designing, learning, and applying Teacher-Student architectures on various distillation objectives. 

\end{abstract}

}

\maketitle

\IEEEdisplaynontitleabstractindextext

%
\IEEEpeerreviewmaketitle


\section{Introduction}

\IEEEPARstart{D}{eep} neural networks (DNNs) have witnessed much success in several fields, such as Computer vision~\cite{minaee2021image} (CV), Communication systems~\cite{erpek2020deep}, and Natural language processing (NLP)~\cite{otter2020survey}, etc. 
Specifically, to satisfy the robust performance in large-scale tasks, DNNs are generally over-parameterized with complex architectures. 
However, such cumbersome models, meanwhile, need a large amount of training time and bring large computational costs, which pose significant challenges to deploying these models on edge devices and in real-time systems.  

To accelerate the training process, Hinton et al.~\cite{hinton2015distilling} first propose \textit{Knowledge Distillation (KD)} technique for training lightweight models to achieve comparable performance to deep models, which is achieved through compressing the informative knowledge from a large and computationally expensive model (i.e., teacher model) to a small and computationally efficient model (i.e., student model).
With such the Teacher-Student architecture, the student model can be trained under the supervision of the teacher model.
During the training of the student model, the student model not only should predict ground truth labels as closely as possible but also should match softened label distributions of the teacher model. 
Consequently, the compressed student model is able to obtain comparable performance to the cumbersome teacher model and is computational-efficiently deployed in real-time applications and edge devices.


In addition to \textit{knowledge compression}, Teacher-Student architectures, meanwhile, are effectively and widely embraced on the other KD objectives, including \textit{knowledge expansion, knowledge adaptation, and knowledge enhancement}. 
With the help of Teacher-Student architectures, we are able to achieve multiple distillation objectives through effective and generalized student networks. 
In \textit{knowledge expansion}, with the stronger model capacity and the complicated learning tasks, student networks can learn the extended knowledge from teacher networks, so that students are able to demonstrate better performance and generalizability over teachers in more complicated tasks~\cite{xie2020self, sohn2020simple, wang2021data}. 
To achieve the objective of \textit{knowledge adaptation}, student networks can be trained on one or multiple target domains, with the adapted knowledge of teacher networks built on source domains~\cite{matiisen2019teacher, li2019bidirectional}.
In \textit{knowledge enhancement}, student networks can learn more general feature representations under the supervision of specialized teacher networks, so that such general student networks can be effectively generalized in multiple tasks~\cite{ghiasi2021multi, yang2022cross}.

\begin{table*}[t]
    \renewcommand{\arraystretch}{1.10}
    \newcommand{\tabincell}[2]{\begin{tabular}{@{}#1@{}}#2\end{tabular}}
    \centering
    \caption{The comparison between the existing KD surveys~\cite{gouKnowledgeDistillationSurvey2021, wang2021knowledge, alkhulaifi2021knowledge} and this survey.}
    \label{tab: survey comparison}
    \begin{tabular}{c|c|c|c|c}
    \toprule
    Work & \citet{gouKnowledgeDistillationSurvey2021} & Wang et al.~\cite{wang2021knowledge} & \citet{alkhulaifi2021knowledge} & \textbf{Our survey} \\
    \midrule
    \tabincell{c}{Distillation \\ objective} & Knowledge compression & Knowledge compression & Knowledge compression & \tabincell{c}{Knowledge compression \\ Knowledge expansion \\ Knowledge adaptation \\ Knowledge enhancement} \\
    \midrule
    \tabincell{c}{Knowledge \\ representation} & \tabincell{c}{Response, Intermediate, \\ Relation} & \tabincell{c}{Response, Intermediate, \\ Mutual information} & \tabincell{c}{Response, Intermediate} & \tabincell{c}{Response, Intermediate, \\ Relation, Mutual information} \\
    \midrule
    \tabincell{c}{Knowledge \\ optimization} & \xmark & \xmark & \xmark & \tabincell{c}{Huber loss, Triplet loss, \\ Information maximization, etc.} \\
    \midrule
    \tabincell{c}{Application \\ purpose} & Classification, Recognition & Classification, Recognition & Classification, Recognition & \tabincell{c}{Classification, Recognition, \\ Generation, Ranking, \\ Regression}\\
    \midrule
    \end{tabular}
\end{table*}

With the recent advancements in Teacher-Student architectures, some studies have summarized the recent progress of various distillation techniques with Teacher-Student architectures.
Specifically, ~\citet{gouKnowledgeDistillationSurvey2021} present a comprehensive KD survey mainly from the following perspectives: knowledge representations, distillation schemes and algorithms. 
Wang et al.~\cite{wang2021knowledge} provide a systematic overview and insight into model compression with Teacher-Student architectures in the vision field. 
Alkhulaifi et al.~\cite{alkhulaifi2021knowledge} summarize multiple metrics to evaluate distillation methods in terms of reduction in model size and performance.

However, existing KD surveys~\cite{gouKnowledgeDistillationSurvey2021, wang2021knowledge, alkhulaifi2021knowledge} mainly focus on the Teacher-Student architectures for the objective of knowledge compression, thereby calling for a comprehensive review of all the distillation objectives. 
Moreover, knowledge types can be summarized into three categories: \textit{response-based, intermediate, relation-based, and mutual information-based representations}. The knowledge optimization objectives may vary depending on the specific knowledge representation. 
However, existing surveys~\cite{gouKnowledgeDistillationSurvey2021, wang2021knowledge, alkhulaifi2021knowledge} only provide a review of, at most, three knowledge representations, lacking a comprehensive introduction to knowledge optimization with various representations.
Besides, the existing works~\cite{gouKnowledgeDistillationSurvey2021, wang2021knowledge, alkhulaifi2021knowledge} primarily explain the applications of Teacher-Student architectures in the domains of visual recognition and NLP, suggesting that other tasks (i.e., generation, ranking, and regression) are also being discussed.

\begin{figure*}[t]
\centering
\includegraphics[width=0.9\textwidth]{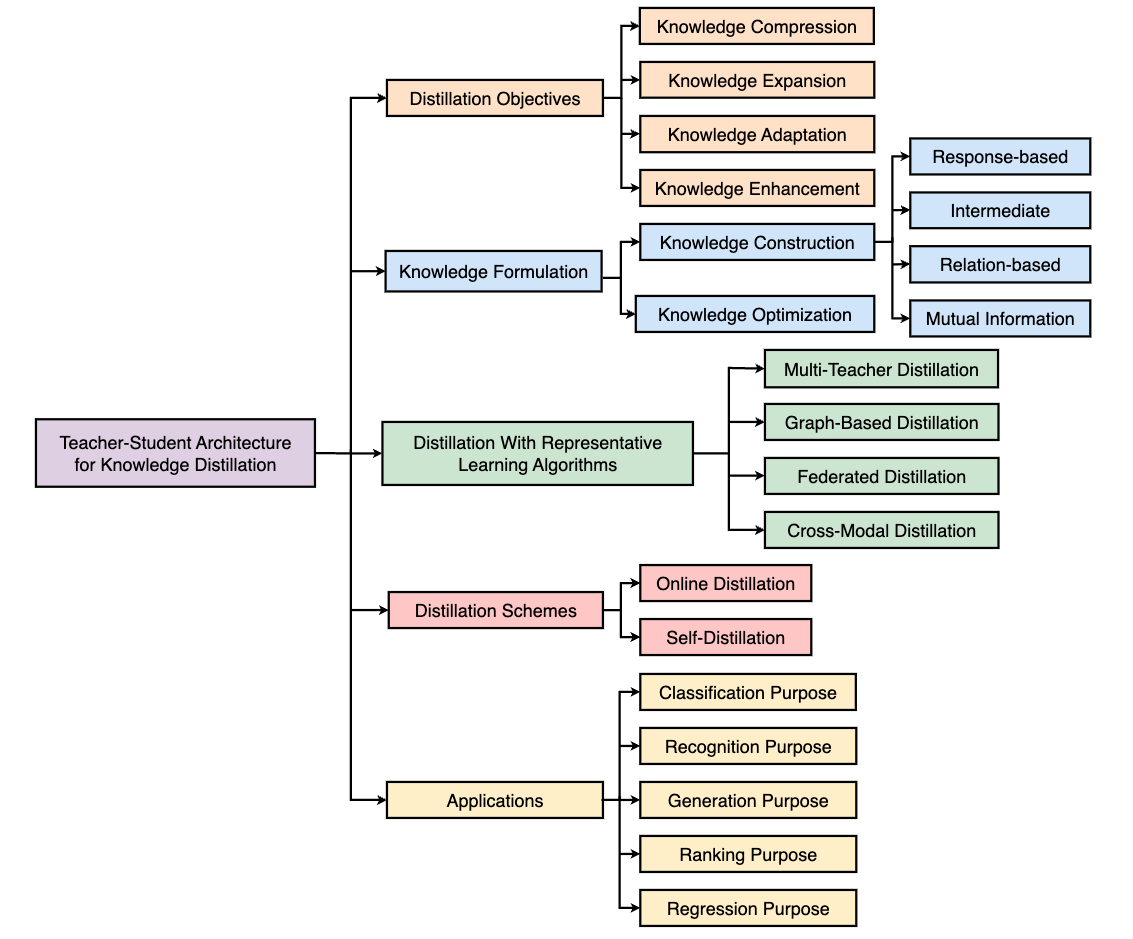}
\centering
\caption{The general taxonomy framework of this survey.}\label{fig:taxonomy}
\vspace{-5mm}
\end{figure*}

To this end, this survey provides a comprehensive and insightful guideline about the Teacher-Student architectures for KD. 
As shown in Fig.~\ref{fig:taxonomy}, the general taxonomy framework of our survey, this survey first discusses Teacher-Student architectures on multiple KD objectives, including \textit{knowledge compression, knowledge expansion, knowledge adaptation, and knowledge enhancement}. 
This survey provides a detailed overview of multiple knowledge representations (i.e., \textit{response-based, intermediate, relation-based, and mutual information-based representations}), and explores the knowledge optimization objectives associated with each specific representation. 
Moreover, we systematically summarize Teacher-Student architectures with multiple representative learning algorithms (i.e., \textit{multi-teacher, graph-based, federated, and cross-modal distillation}), while introducing online distillation and self-distillation schemes under the framework of Teacher-Student architectures. 
The latest applications of Teacher-Student architectures are also presented, providing insights from various perspectives:  \textit{classification, recognition, generation, ranking, and regression purposes}.
Finally, we investigate the potential research directions of KD on Teacher-Student architecture design, knowledge quality, and theoretical studies of regression-based learning, respectively.

Table~\ref{tab: survey comparison} compares the prior works~\cite{gouKnowledgeDistillationSurvey2021, wang2021knowledge, alkhulaifi2021knowledge} with our survey, summarizing the \textbf{main contributions} of this survey:

\begin{itemize}
    \item We introduce a comprehensive review of Teacher-Student architectures for multiple distillation objectives, including \textit{knowledge compression, knowledge expansion, knowledge adaptation, and knowledge enhancement}.  
    \item We provide an insightful review of multiple knowledge representations and optimization objectives associated with each specific representation. 
    \item We summarize recent applications of Teacher-Student architectures across various purposes, including \textit{classification, recognition, generation, ranking, and regression}. 
    \item We discuss promising research directions on KD, including Teacher-Student architecture design, knowledge quality, and theoretical studies in regression-based learning. 
\end{itemize}

The rest of the survey is organized as follows: Section~\ref{sec:learning objective} describes Teacher-Student architectures across multiple distillation objectives. Section~\ref{sec: knowledge formulation} introduces knowledge representations and optimization objectives. Section~\ref{sec:architecture} and Section~\ref{sec:learning scheme} discuss Teacher-Student architectures with representative learning algorithms and distillation schemes, respectively. 
Section~\ref{sec:application} summarizes recent applications of Teacher-Student architectures across various purposes. The future works and conclusions are eventually drawn in Section~\ref{sec:future work} and Section~\ref{sec:conclusion}, respectively.

\section{Distillation Objectives}\label{sec:learning objective}
In this section, we will introduce Teacher-Student architectures for multiple distillation objectives, including knowledge compression, knowledge enhancement, knowledge adaptation, and knowledge expansion.


\subsection{Knowledge Compression}


Knowledge compression focuses on training a student model, using predictions from a larger-sized teacher model. The purpose of knowledge compression is to achieve a compact student model while maintaining comparable or slightly reduced performance with a teacher model. 

Hinton et al.~\cite{hinton2015distilling} first propose to distill knowledge from multiple models to a single student model for the task of model compression and transfer learning. Tang et al.~\cite{tang2019distilling} compress BERT~\cite{devlin2018bert} to a much light-weight Bi-LSTM~\cite{huang2015bidirectional} for the task of natural language processing. Romero et al.~\cite{romero2014fitnets} suggest that the success of deep neural nets is largely attributed to the deep hierarch. Thus they propose to compress wide (a large number of neurons in each layer) and deep teacher models into much narrower (fewer neurons in each layer) and deeper student models. Yim et al.~\cite{yim2017gift}, design the architectures of students and teachers as a $N$-part module, where each module contains various numbers of convolutional layers. Student models generally have a simpler design, and the task for students is to learn each layer output of the teacher.

In addition, Wang et al.~\cite{wang2018kdgan} argue that the vanilla KD~\cite{hinton2015distilling} is hard for students to learn all knowledge from teachers, and thus students normally show worse performance compared to the teacher. 
Hence, the authors~\cite{wang2018kdgan} adopt Generative adversarial networks (GANs)~\cite{goodfellow2014generative} to simulate the compression process. The generator (student model with fewer parameters) learns the distribution of the data, whereas the discriminator (teacher model with more parameters) learns to differentiate if the input is from a student or real.
Tang et al.~\cite{tang2018ranking} propose to distill complicated models in information retrieval or recommendation systems where the model predicts the top K relevant information when a query is given. Zhang et al.~\cite{zhang2018deep} utilize multiple students (without teachers) to learn from each other during training. Furlane et al.~\cite{furlanello2018born} propose a novel ensemble learning approach, where students and teachers share the same architecture. The $N_{th}$ student is responsible to train the $N+1_{th}$ student. Predictions are averaged in the end.

\begin{figure*}
\begin{subfigure}{0.23\linewidth}
  \centering
  \includegraphics[width=\linewidth]{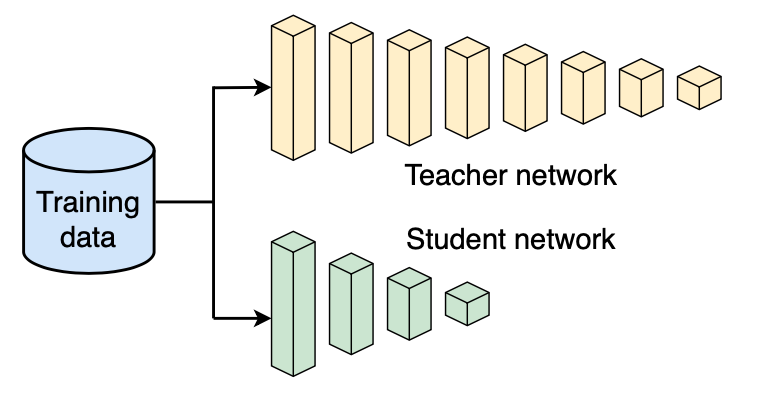}
  \caption{Knowledge compression.}
\end{subfigure}\hfill
\begin{subfigure}{0.25\linewidth}
  \centering
  \includegraphics[width=\linewidth]{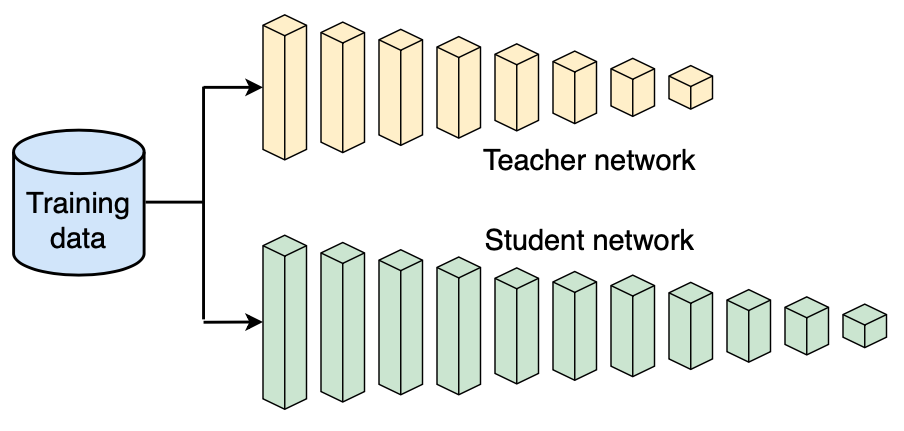}
  \caption{Knowledge expansion.}
\end{subfigure}\hfill
\begin{subfigure}{0.26\linewidth}
  \centering
  \includegraphics[width=\linewidth]{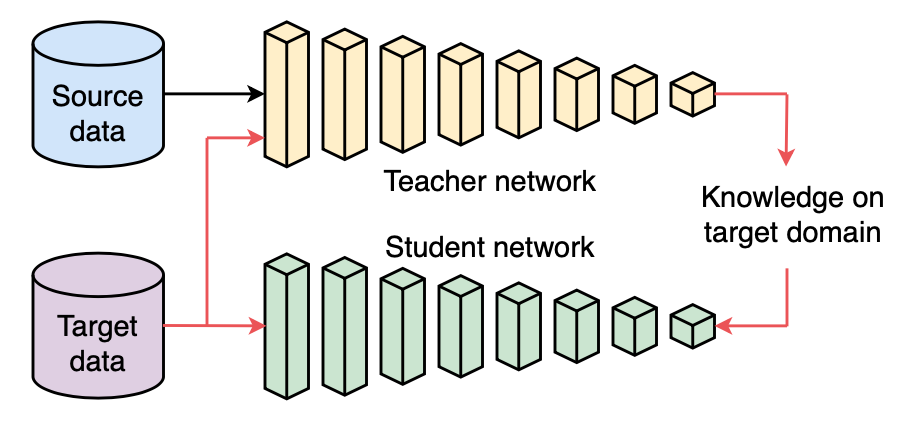}
  \caption{Knowledge adaptation.}
\end{subfigure}\hfill
\begin{subfigure}{0.25\linewidth}
  \centering
  \includegraphics[width=0.75\linewidth]{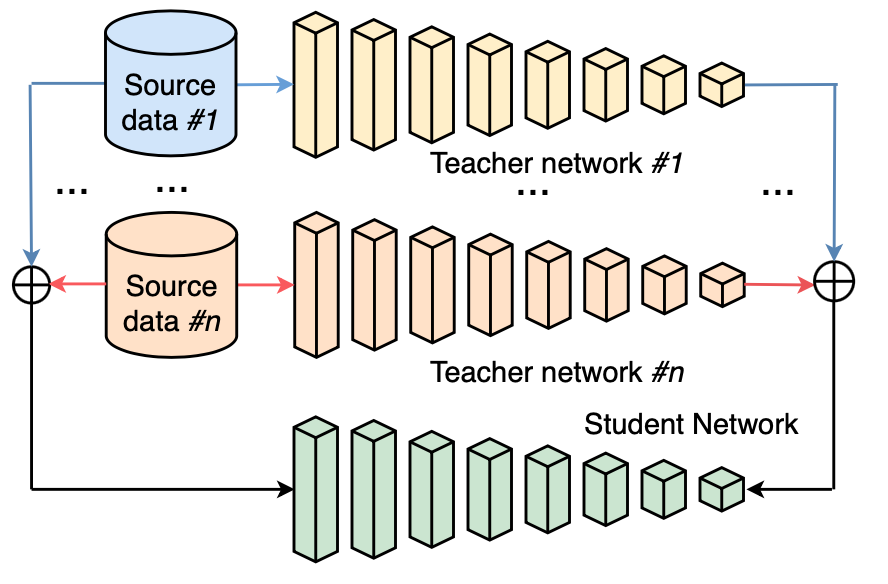}
  \caption{Knowledge enhancement.}
\end{subfigure}\hfill
\caption{Teacher-Student architectures with the various distillation objectives.}
\label{fig: distillation objectives}
\vspace{-5mm}
\end{figure*}

\subsection{Knowledge Expansion}


Instead of compressing knowledge from larger teacher networks into smaller student networks, knowledge expansion focuses on building student networks with stronger learning capacity and performance than teacher networks. 
The knowledge can be expanded in multiple approaches, such as increasing the size of the student network, data augmentation, and introducing random noise to student networks, among others.  
These strategies are able to enhance the robustness and generalizability of the student network, contributing to their improved performance in diverse scenarios.

Specifically, Xie et al.~\cite{xie2020self} initially propose the concept of knowledge expansion, which aims to use a teacher model to train a student model with a larger capacity of parameters. Xie et al.~\cite{xie2020self} initially train the teacher model on ImageNet~\cite{deng2009imagenet}, then incorporate a privately collected unlabelled dataset, and use the teacher model's prediction as pseudo ground truth to train a larger student model. With the enhanced data augmentation and the additional self-labeled data, the larger student model can outperform the teacher model through an iterative self-training strategy.
\citet{sohn2020simple} adopt a similar training strategy for the task of object detection, which utilizes the pre-trained teacher model to generate pseudo labels for the unlabelled images. The student model is trained with strong data augmentation on the pseudo-labeled dataset. Wang et al.~\cite{wang2021data} argue that forcing student models to learn the noisy pseudo labels from teachers may cause noisy label overfitting. They instead propose a curriculum learning strategy, where the student models learn easy samples first, then proceed to hard samples. The criteria for easiness are based on the confidence score from the region proposal network.

\subsection{Knowledge Adaptation}


To fulfill the objective of knowledge adaptation, we are expected to train well-generalized student networks on one or multiple unknown target domains by leveraging the adapted knowledge from teacher networks constructed on similar source domains.

In the work of~\cite{matiisen2019teacher}, teacher models monitor students' learning progress and decide which task each student should be trained on. In general, a student should be trained on the task where he gains the most performance improvement. To prevent Catastrophic Forgetting~\cite{kirkpatrick2017overcoming}, students should be trained on tasks where the performance drops. 
\citet{tsai2018learning} adopt a novel teacher-student learning scheme by incorporating GANs~\cite{goodfellow2014generative} into the training. The authors~\cite{tsai2018learning} perform supervised segmentation tasks on the source images with labels, and then use the trained model to generate predictions for both source and target images. Lastly, they feed both predictions to a teacher model (i.e., discriminator). If a teacher model is incapable of differentiating which domain the prediction is from, the trained model has successfully transferred the knowledge from the source domain to the target domain.
\citet{hoffman2018cycada} introduce CycleGAN~\cite{zhu2017unpaired} to align source domain images with the target domain in an unsupervised approach. Then, they further use the aligned source images along with target images to join the adversarial training similar to~\cite{tsai2018learning} where the discriminator is treated as the teacher network.
Furthermore, ~\citet{li2019bidirectional} extend the work of~\cite{hoffman2018cycada} through a bi-directional learning scheme, where the adversarial training can be in turn used to promote the training of CycleGAN~\cite{zhu2017unpaired}. 

\subsection{Knowledge Enhancement}


Despite the effectiveness of the Teacher-Student training paradigm in various domains, adapting Teacher-Student architectures to a multi-task context still remains challenging. Therefore, the objective of knowledge enhancement is to train student networks with more generalized feature representations under the guidance of specialized teacher networks. This enables the general student networks to consistently exhibit robust performance across multiple tasks.

Ghiasi et al~\cite{ghiasi2021multi} propose a multi-task self-training (MuST) strategy which uses multiple independent teacher models to train one multi-task student model. In particular, they adopt four teacher models, each responsible for classification, detection, segmentation, and depth estimation on four different datasets. After training, these teacher models are used to generate four types of pseudo labels for much larger datasets. The student model is then trained on the dataset with four types of pseudo labels.
\citet{yang2022cross} propose a cross-task framework consisting of three modules: 1) task augmentation: ranking fine-grained cross-tasks through an auxiliary loss. 2) distillation: sharing ranked knowledge representation across tasks to enforce consistency. 3) Teacher-Student training: end-to-end training to improve the generalized representation of distillation. 

\subsection{Comparison Analysis}

In summary, we compare the aforementioned distillation objectives as shown in Fig.~\ref{fig: distillation objectives} and \textbf{Table 1 of the supplementary material}.
Specifically, Fig.~\ref{fig: distillation objectives} shows Teacher-Student architecture with each specific objective. 
\textit{Knowledge compression} aims to train a compact student network (with fewer parameters) by distilling the informative knowledge from a large teacher network. The student network can significantly reduce computational costs compared to teacher networks, while achieving comparable performance.
In \textit{knowledge expansion}, student networks are expected to surpass the performance and generalization capabilities of teacher networks by enriching the distilled knowledge from teachers. To expand knowledge, student networks can be trained in a more complex environment, such as increasing the network size, and introducing random noise, among others. Through these approaches, student networks can exhibit robust performance across diverse and intricate  scenarios, ultimately outperforming their teacher networks. 
\textit{Knowledge adaptation} focuses on training well-generalized student networks on one unknown target domain, with the adapted knowledge from teacher networks built on similar source domains. 
Given that each teacher network is constructed in a specific task-oriented environment, \textit{knowledge enhancement} aims to train one generalized student network that simultaneously exhibits strong performance across multiple tasks, which is achieved by integrating task-specific knowledge from each unique teacher network.

In addition, \textbf{Table 1 of the supplementary material} compares the distillation objectives from multiple perspectives: knowledge representation, distillation scheme, student size (over teacher), and teacher number. It is observed that knowledge can be constructed with various representations: response (e.g., logits~\cite{hinton2015distilling}), intermediate (e.g., feature maps~\cite{romero2014fitnets}), relation (e.g., sample relations~\cite{Park_2019_CVPR}), and mutual information (e.g., Kernel-based information flow~\cite{Passalis_2020_CVPR}), which will be detailedly discussed in Section~\ref{sec: knowledge formulation}. 
The student networks can be trained using diverse distillation schemes among all the objectives. In self-distillation, the roles of student and teacher networks are dynamic during the iterative learning process, which indicates that student and teacher networks can be exchanged or student networks are able to learn the knowledge from themselves (without teacher networks). The detailed review of distillation schemes will be provided in Section~\ref{sec:learning scheme}. 
Furthermore, since the objective of \textit{knowledge compression} is to construct compact student networks, it is important to note that the size of student networks is smaller than that of teacher networks. 
The student networks can be designed with a larger or identical size to the teacher networks in the works~\cite{xie2020self, sohn2020simple, wang2021data}, providing feasible approaches for \textit{knowlegde expansion}. 
To obtain well-generalized student networks in \textit{knowledge enhancement}, multiple teacher networks are constructed across diverse scenarios. The introduction of multi-teacher distillation will be summarized in Section~\ref{sec:architecture}.

We further report the classification performance and computational efficiency of different methods across the multiple distillation objectives in \textbf{Table 2 and Table 3 of the supplementary material}, respectively.
By acquiring the distilled knowledge from teacher networks, lightweight student networks are trained in a computationally efficient manner, while consistently demonstrating effective performance across various datasets for all the distillation objectives. 
Notably, CIFAR100 and CIFAR10 datasets~\cite{krizhevsky2009learning} are commonly employed for evaluation purposes. 
Compared to the vanilla KD~\cite{hinton2015distilling}, which solely relies on logits as knowledge, FitNet~\cite{romero2014fitnets} introduces intermediate feature maps as additional knowledge, leading to an improved classification accuracy of $0.26\%$. This suggests the advantage of incorporating intermediate feature maps in knowledge compression.
The computational cost can be significantly reduced, particularly when multiple teacher networks are involved. 
In both the hetero-architecture and homo-architecture designs of teacher networks, student networks can consistently exhibit strong performance across all the objectives.

\section{Knowledge Formulation}\label{sec: knowledge formulation}
In this section, we will not only explain multiple knowledge representations during the knowledge construction, but also summarize knowledge optimization objectives associated with each specific knowledge representation.

\subsection{Knowledge Construction}

\subsubsection{Response-Based Knowledge}

The most straightforward knowledge representation is response-based knowledge, which is constructed through the final prediction of teacher networks. As explained in the work of Hinton et al.~\cite{hinton2015distilling}, such final predictions of teachers contain informative dark knowledge, which can be generalized from a complicated teacher network to a small student network. 

The response-based knowledge can be specifically formulated for different application tasks, thereby incorporating task-specific auxiliary knowledge, such as image classification, pose estimation, object detection, and speech recognition, among others. 
For example, the most common response-based knowledge is represented as soft targets in image classification~\cite{hinton2015distilling, xie2020self, stanton2021does, Beyer_2022_CVPR, zhu2018knowledge, Zhao_2022_CVPR}. Denoting $\mathbf{z}_i^T$ as the logit of teacher network for the $i$-th class, soft target is a vector of the prediction probabilities of all classes, which is calculated through a \texttt{Softmax} function as follows:
\begin{equation}
    P_{i}^{T}(\mathbf{z}_i^T, \tau)=\frac{\texttt{exp}(\mathbf{z}_i^T/\tau)}{\sum_{j}\texttt{exp}(\mathbf{z}_j^T/\tau)}, 
\end{equation}
where $P_{i}^{T}$ is the soft prediction probability for the $i$-th class, and $\tau$ is a temperature factor controlling the softness of logits. By regressing such soft prediction probabilities of all classes, the student network can be effectively trained with informative supervision from teacher networks. 
Specifically, Zhao et al.~\cite{Zhao_2022_CVPR} further categorize the soft target into two parts: (1) distilled knowledge on target class (i.e., TCKD) concerning the ``difficulty" of training samples, and (2) distilled knowledge on non-target class (i.e., NCKD) explaining the reason why logit distillation works. The effectiveness and flexibility of the distillation process can be improved through a weighted sum of the two parts.

Considering that the soft-target-based knowledge cannot represent the structural information in 2D image space, confidence maps~\cite{Zhang_2019_CVPR, Zhao_2018_CVPR, Nie_2019_ICCV} and heatmaps~\cite{li2021online, Wang_2021_CVPR} are commonly considered as the knowledge to be transferred from teacher models in human pose estimation.
Zhang et al.~\cite{Zhang_2019_CVPR} generate a confidence map $\mathbf{m}_{k}$ for each single joint $k$ by centering a Gaussian kernel around the relevant pixel position $(x_k, y_k)$ as follows:
\begin{equation}
    \mathbf{m}_{k}(x,y)=\frac{1}{2\pi\sigma^{2}}\exp\Big(\frac{-[(x-x_{k})^{2}+(y-y_{k})^{2}]}{2\sigma^{2}}\Big), 
\end{equation}
where $(x, y)$ denotes a pixel location and $\sigma$ is s pre-fixed spatial variance. With such the structural knowledge, the student network is trained to align with the confidence map predicted by the pre-trained teacher network. 
In the work of Li et al.~\cite{li2021online}, pose estimation predicts a heatmap for each human anatomical keypoint, indicating the keypoint locations as Gaussian peaks. Since pixel values on the heatmap represent the probabilities of each pixel in the keypoint, Li et al.~\cite{li2021online} further transfer the pose structural knowledge by minimizing the divergence between the probabilities of pixels generated from the teacher and student network.

In object detection, bounding box regression is one powerful method to adjust the location and size of input proposals, which is critical to ensure precise detection performance~\cite{dai2016r, ren2015faster, 7546875}. Thus, the bounding box is commonly represented as response-based knowledge, which is considered as one primary target for students to learn from teachers~\cite{chen2017learning, Chawla_2021_WACV, Shmelkov_2017_ICCV, Zheng_2022_CVPR}. 
For example, in addition to the classification output, Chen et al.~\cite{chen2017learning} also encourage the student network to mimic the predicted bounding box of the teacher network, serving as an upper bound for the student's performance. The teacher could potentially provide misleading guidance to the student, due to the unbounded nature of real-valued regression outputs. Thus, the student is only allowed to match the regression outputs when the error of the student is significantly larger than the teacher. 

Besides, automatic speech recognition suffers from performance degradation when a well-trained acoustic model is applied in a new domain~\cite{6732927}. To tackle this issue, teacher-student architectures incorporate domain adaptation to adapt a source-domain acoustic model to target-domain speech, which transfers the source-domain knowledge for the student built specifically on the target domain~\cite{meng2019conditional, li2018developing, chebotar2016distilling, 8683710, 8639566}. 
In the work of Li et al.~\cite{li2018developing}, the close-talk data is the source data $x_{src}$ inputting the teacher network while the far-field data is the target data $x_{tgt}$ (i.e, the input of the student network). The student network is trained to minimize the KL divergence between the posterior distributions of the teacher $P_{T}(s\vert x_{src})$ and student $P_{S}(s\vert x_{tgt})$, which enables the student to achieve effective generalization on both known and novel scenarios.
Meng et al.~\cite{meng2019conditional} further introduce a conditioned learning scheme, in which the student network can criticize the response-based knowledge provided by the teacher network. The student considers the posterior distribution of the teacher as the knowledge when the teacher's prediction is correct, and otherwise the ground truth label is regarded as the learning target by the student.

\subsubsection{Intermediate Knowledge}


DNNs exhibit the strong capability to learn abstract and invariant feature representations at intermediate layers~\cite{6472238}, which motivates the introduction of intermediate-level representations as extended knowledge to guide the training of the student network. 

FitNet~\cite{romero2014fitnets} is the first work to construct intermediate knowledge through a hint layer, where the knowledge is the intermediate output of such layer in a teacher network. However, a linear mapping $\beta$ is introduced to unify different dimensions as a result of the dimension difference between the intermediate outputs of the teacher and student networks. For knowledge distillation, the squared Euclidean distance between hidden activations is employed. The knowledge used by the authors~\cite{romero2014fitnets}, such as the input feature and output feature of a residual block, is a Gram matrix of feature mappings from two separate layers. The student network's weight initialization for future optimization is based on the information that has been condensed. The performance of the student network in various tasks, according to the authors, can be enhanced by the knowledge that has been condensed.
Inspired by FitNet~\cite{romero2014fitnets}, numerous studies encourage student networks to not only learn response-based knowledge, but also to receive training supervision with intermediate-layer feature representations from teacher networks~\cite{Nie_2019_ICCV, chen2017learning}. 
Besides, activation and gradient attention maps can be represented as the intermediate knowledge in attention-based works \cite{zagoruykoPayingMoreAttention2017, kimParaphrasingComplexNetwork2018, heoKnowledgeTransferDistillation2019}. The normalized sum of absolute or squared value mapping determines the attention maps.
\citet{zagoruykoPayingMoreAttention2017} demonstrate that these attention maps include useful information that can help convolutional neural network architectures perform better.
In order to create L$_2$-normalized feature embeddings of teacher network as knowledge, Kim et al. \cite{kimParaphrasingComplexNetwork2018} suggest an encoder with a convolutional structure. It is encouraged for the student network to generate the same compressed representation as the teacher network.
As a classification problem heavily depends on the construction of decision boundaries among classes, Heo et al. \cite{heoKnowledgeTransferDistillation2019} employ the pattern of active outputs as knowledge. The teacher network retains a lot of valuable information because the pattern is extracted before the activation function. Prior to the classification training, the student network is initialized using the condensed knowledge.

Based on the observation that pair-wise activations with semantically comparable inputs frequently produce similar output patterns, the author of \cite{heoKnowledgeTransferDistillation2019} employs the L$_2$-normalized outer products of pair-wise activations as knowledge. As a result, while processing the same input pairs, the student network will result in output patterns that are similar to (or different from) those of the trained teacher network.
Through the teacher network, they also create feature maps using a collection of samples. And consider the structural knowledge contained within those feature maps. The average Euclidean distance and cosine similarity between grouped sample outputs define the structural information, or distance metric. Before the knowledge distillation process, both the teacher network and the student network have been pre-trained.
Anchor loss in metric learning is used in the work of DarkRank \cite{chenDarkrankAcceleratingDeep2018} to formulate the information. The anchor loss increases the distance between classes while reducing the distance within them. These patterns of intra- and inter-class distance are employed as knowledge. The hidden activations are also encoded by the authors using the embedding layer. The knowledge distillation process is optimized using the KL divergence and MLE.
A hybrid structure to direct the student network is revealed by learning from several teacher networks. The authors utilize triplet loss to extract intra- and inter-class information as one knowledge and average the softened outputs of several teacher networks as another knowledge. The knowledge distillation loss is totaled as the ultimate loss according to the voting approach of multiple teachers.

\subsubsection{Relation-Based Knowledge}

In addition to response-based and intermediate knowledge, the correlations among samples are also valuable knowledge for enhancing the performance of the student network, since such relation-based knowledge can capture the structural information in the data embedding space~\cite{huang2022knowledge, Park_2019_CVPR, Peng_2019_ICCV, Yang_2022_CVPR, Liu_2019_CVPR}. Specifically, preserving the relations of final predictions from the teacher network is sufficient and effective.  the correlations among samples are also valuable knowledge for enhancing the performance of the student network, since such relation-based knowledge can capture the structural information in the data embedding space. 
Instead of distilling the exact prediction probability, \citet{huang2022knowledge} leverage the Person correlation coefficient~\cite{pearson1896vii} to transfer the inter-class relations (i.e., the relative ranking of predictions on each sample) and the intra-class relations (i.e., the relative ranking of predictions on each class) from the teacher to the student. 
Park et al.~\cite{Park_2019_CVPR} explore the relations among training samples through distance-wise $\psi_{D}(t_i,t_j)$ and angle-wise $\psi_{A}(t_i,t_j,t_k)$ functions as follows:
\begin{align}
    \psi_{D}(t_i,t_j)&=\frac{1}{\mu}\Arrowvert t_{i}-t_{j}\Arrowvert_{2}, \\
    \psi_{A}(t_i,t_j,t_k)&=\texttt{cos}\measuredangle t_it_jt_k=\langle \mathbf{e}^{ij}, \mathbf{e}^{kj} \rangle,
\end{align}
where $t_i$ is the final prediction output of the teacher network on one sample, and $\mu$ is a normalization factor. $\psi_{D}(t_i,t_j)$ calculates the Euclidean distance between two samples in the output representation space. 
Given a triplet of samples, $\psi_{A}(t_i,t_j,t_k)$ measures the angular relation formed by two edges as follows:
\begin{equation}
    \mathbf{e}^{ij}=\frac{t_i-t_j}{\Arrowvert t_{i}-t_{j}\Arrowvert_{2}}, \mathbf{e}^{kj}=\frac{t_k-t_j}{\Arrowvert t_{k}-t_{j}\Arrowvert_{2}}. 
\end{equation}

In addition to the final prediction output space, relation-based knowledge can also be discovered by measuring the correlations among samples in the intermediate feature space, which promotes a comprehensive understanding of the dependencies within samples. 
Peng et al.~\cite{Peng_2019_ICCV} introduce the Gaussian RBF kernel-based function to capture the high-order correlation between samples as follows:
\begin{align}
    k(f_i,f_j) &= \texttt{exp}(-\gamma\Arrowvert f_{i}-f_{j}\Arrowvert^{2}), \\
    &\approx \sum_{p=0}^{P}\texttt{exp}(-2\gamma)\frac{(2\gamma)^{p}}{p!}(f_{i}\cdot f_{j}^{\top})^{p},
\end{align}
which can be approximated by $P$-order Taylor series. $f_i$ and $f_j$ are the feature representations captured from the teacher network on two samples, and $\gamma$ is a tunable parameter. Furthermore, Yang et al.~\cite{Yang_2022_CVPR} capture the cross-image pair-wise correlations among pixels, by the pixel similarity matrix (i.e., element-wise dot product computed by the Bilinear pooling~\cite{Lin_2015_ICCV}) between the anchors and contrastive embeddings for the teacher network. 
Liu et al.~\cite{Liu_2019_CVPR} model the instance relation graph as the transferred knowledge, in which each vertex represents one instance feature representation at the intermediate layer and each edge is defined as the Euclidean distance between the intermediate features of two instances.

\subsubsection{Mutual Information-Based Knowledge}

From the perspective of information theory, knowledge transfer can be formulated as the maximization of the mutual information between teacher and student networks. Specifically, the mutual information between two random variables $(t, s)$ can be formulated as follows:
\begin{align}
    I(t; s) &=H(t)-H(t|s),\\
    &= -\mathbb{E}_{t}[\log p(t)]+\mathbb{E}_{t,s}[\log p(t|s)],
\end{align}
where the entropy $H(t)$ and the conditioned entropy $H(t|s)$ are derived from the joint distribution $p(t,s)$. To this end, $I(t; s)$ can indicate the uncertainty reduction in the knowledge of the teacher network encoded in the $t$-th layer when the $s$-th layer of the student network is known. 

Considering that the true distribution $p(t,s)$ could be unknown, Ahn et al.~\cite{Ahn_2019_CVPR} propose a variational lower bound for $I(t;s)$ by introducing a variational distribution $q(t|s)$ approximating $p(t|s)$:
\begin{align}
    I(t; s) &=H(t)+\mathbb{E}_{t,s}[\log p(t|s)], \\
    &=H(t)+\mathbb{E}_{t,s}[\log q(t|s)]+\mathbb{E}_{s}[D_{KL}(p(t|s)\parallel q(t|s))], \\
    &\geq H(t)+\mathbb{E}_{t,s}[\log q(t|s)], \label{eqn:mutual information}
\end{align}
which aims to maximize such the variational lower bound through the variational information maximization scheme~\cite{barber2004algorithm}. The Gaussian distribution with heteroscedastic mean $\mu(\cdot)$ and homoscedastic variance $\sigma$ can be modelled as $q(t|s)$, where the relevant parameters are specified based on the intermediate and logit layers of the teacher network. 
To better retain sufficient and task-relevant knowledge, Tian et al.~\cite{Tian_2021_CVPR} utilize the information bottleneck to produce highly-represented encodings $z$ \textit{w.r.t} ground truth labels $y$, namely the sufficiency $I(z;y)$ of $z$ for $y$. Through equivalently maximizing the subtraction of conditional entropy $H(y|z)$ and $H(y|v)$, the optimal representation can be achieved with minimized superfluous information (i.e., task-irrelevant knowledge). 
To capture the structural knowledge (i.e., high-order relations among sample representations), Zhu et al.~\cite{Zhu_2021_CVPR} and Tian et al.~\cite{tian2019contrastive} maximize the low bound of mutual information between the anchor-teacher relation and the anchor-student relation, thereby providing a feasible solution in the contrastive learning. 
Shrivastava et al.~\cite{Shrivastava_2023_CVPR} propose three mutual information maximization objectives between the teacher and student networks: (1) global mutual information between the final representations, (2) local mutual information between region-specific vectors extracted from an intermediate representation of the student network and the final representation of the teacher network, and (3) feature mutual information between region-consistent local vectors extracted from intermediate representations.

In addition, Passalis et al.~\cite{Passalis_2020_CVPR, Passalis_2018_ECCV} utilize the Quadratic mutual information (QMI)~\cite{ torkkola2003feature}, derived by replacing the KL divergence with a quadratic divergence measure, to define the information of teacher and student networks. 
Specifically, this paper~\cite{Passalis_2020_CVPR} models the information flow path of the teacher network as the progression of QMI between each intermediate representation and training targets. The QMI is further expressed as the conditional probability distribution:
\begin{equation}
    p_{i|j}^{(t,l_{t})}=\frac{K(\mathbf{x}_{i}^{(l_t)}, \mathbf{x}_{j}^{(l_{t})})}{\sum_{i=1,i\neq j}^{N}K(\mathbf{x}_{i}^{(l_t)}, \mathbf{x}_{j}^{(l_{t})})} \in [0,1],
\end{equation}
where $\mathbf{x}_{i}^{(l_t)}$ denotes the representation extracted from the $l$-th layer of the teacher network on the $i$-th sample, and $K(\cdot)$ is the consine and T-student kernel functions. 
Besides, Passalis et al.~\cite{Passalis_2018_ECCV} extend QMI to information potentials using three probabilities $V_{IN}^{(t)}, V_{ALL}^{(t)}$, and $ V_{BTW}^{(t)}$, representing the relations among different samples. $V_{IN}$ denotes the in-class relations, and $V_{ALL}$ denotes the relations between all samples, while $V_{BTW}$ is the relation of each class against all the other samples.

\subsection{Knowledge Optimization}

In knowledge optimization, the objective function of KD can consist of three parts: regular cross-entropy ($\mathcal{L}_{CE}$) term, Kullback–Leibler (KL) divergence ($\mathcal{L}_{KL}$) term, and distance/angle-wise ($\mathcal{L}_{D}$) term.
The generalized form of the objective function of knowledge distillation is:
\begin{equation}
    \mathcal{L}_{K D}=\lambda_1\mathcal{L}_{CE}\left(\mathbf{y}_{\text {true}}, {P}_{{S}}\right)+\lambda_2\mathcal{L}_{KL}\left({P}_{{T}}^{\tau}, {P}_{{S}}^{\tau}\right)+\lambda_3\mathcal{L}_{D},
\end{equation}
where $\lambda_1+\lambda_2+\lambda_3=1$. $\mathcal{L}_{CE}$ refers to the cross-entropy loss measuring the difference between the ground truth $\mathbf{y}_{\text {true}}$ and the student prediction $P_{S}$. $\mathcal{L}_{KL}$ refers to KL divergence between the soft logits of teacher $P_T^{\tau}$ and student $P_S^{\tau}$, where $\tau$ is the temperature factor~\cite{romero2014fitnets} of a \texttt{Softmax} function. $\mathcal{L}_{D}$ refers to distance-wise and angle-wise functions to optimize the other response-based knowledge from teacher networks. 

\textbf{Table 4 of the supplementary material} summarizes the optimization objectives with different knowledge representations. In the response-based knowledge, student networks are commonly trained to learn the soft logits from teacher networks and the ground truths by minimizing $\mathcal{L}_{KL}$ and $\mathcal{L}_{CE}$~\cite{hinton2015distilling, zhu2018knowledge, mirzadeh2020improved, li2017large, yuan2020revisiting, guo2020online}. Besides, $\mathcal{L}_{1}$ and $\mathcal{L}_{2}$ distance functions are represented as $\mathcal{L}_{D}$ to optimize the other response-based knowledge (i.e., confidence maps~\cite{Zhang_2019_CVPR}, heatmaps~\cite{li2021online}, and bounding boxes~\cite{Chawla_2021_WACV, Zheng_2022_CVPR}) as follows:
\begin{align}
    \mathcal{L}_{1} &=\left\|\mathbf{d}_{T}-\mathbf{d}_{S}\right\|, \\
    \mathcal{L}_{2} &=\left\|\mathbf{d}_{T}-\mathbf{d}_{S}\right\|^{2}_2,
\end{align}
where $\mathbf{d}_{T}$ and $\mathbf{d}_{S}$ denote the knowledge to be measured from teacher and student networks. The $\mathcal{L}_{1}$ and $\mathcal{L}_{2}$ functions are also utilized to calculate the distance between the feature maps and attention maps from teacher and student networks~\cite{romero2014fitnets, chen2017learning, zagoruykoPayingMoreAttention2017, kimParaphrasingComplexNetwork2018, wangAdversarialLearningPortable2018, zhang2019your}. In addition to $\mathcal{L}_{2}$ function, \citet{aguilarKnowledgeDistillationInternal2020} also minimize the cosine similarity loss to learn the attention maps of the hidden layers from teachers: 
\begin{align}
    \mathcal{L}_{cos} &= 1-\texttt{cos}(\mathbf{d}_{T}, \mathbf{d}_{S}), \\
    & = 1-\frac{\mathbf{d}_{T}\cdot\mathbf{d}_{S}}{\left\|\mathbf{d}_{T}\right\|\left\|\mathbf{d}_{S}\right\|}. 
\end{align}   
To learn the structural knowledge among samples, \citet{Park_2019_CVPR} and \citet{liu2020adaptive} denote the Huber loss as $\mathcal{L}_{D}$:
\begin{equation}
    \mathcal{L}_{Huber}=
    \begin{cases}
        \frac{1}{2}(\mathbf{d}_{T}-\mathbf{d}_{S})^2 &\mbox{for $|\mathbf{d}_{T}-\mathbf{d}_{S}|\leq 1$}, \\
        |\mathbf{d}_{T}-\mathbf{d}_{S}|-\frac{1}{2} &\mbox{otherwise},
    \end{cases}
\end{equation}
where $\mathbf{d}_{T}$ and $\mathbf{d}_{S}$ can be distance-wise and angle-wise metric (e.g., Euclidean distance and cosine similarity) to measure the structural relations among different samples. 
Given a triplet $(\mathbf{x}_{i}, \mathbf{x}_{i}^{+}, \mathbf{x}_{i}^{-})$, where $\mathbf{x}_{i}$ is the anchor point and $\mathbf{x}_{i}^{+}$ has smaller distance $\psi_{D}^{T}(\mathbf{x}_{i}, \mathbf{x}_{i}^{+})$ with $\mathbf{x}_{i}$ than that $\psi_{D}^{T}(\mathbf{x}_{i}, \mathbf{x}_{i}^{-})$ of $\mathbf{x}_{i}^{-}$ according to the predictions from the teacher network, \citet{you2017learning} adopt the triplet loss to learn such the distance-wise sample relations from the teacher to the student:
\begin{equation}
    \mathcal{L}_{triplet}=\texttt{max}(0, \psi_{D}^{T}(\mathbf{x}_{i}, \mathbf{x}_{i}^{+})-\psi_{D}^{T}(\mathbf{x}_{i}, \mathbf{x}_{i}^{-})+\delta), 
\end{equation}
where $\delta>0$ is a pre-defined small number to prevent the trivial solution. 
From the perspective of information-theoretic criteria, the studies of \citet{Ahn_2019_CVPR}, \citet{tian2019contrastive} and \citet{Shrivastava_2023_CVPR} maximize the KL-based mutual information between student and teacher representations according to Eqn.~(\ref{eqn:mutual information}). To generate highly-represented feature embeddings, \citet{Tian_2021_CVPR} conduct variational mutual learning through Jenson-Shannon divergence (JSD) as:  
\begin{equation}
    \mathcal{L}_{JSD}=\mathbb{E}_{v_1,v_2}\mathbb{E}_{z_1,z_2}[D_{JS}[p(y_1|z_1)\parallel p(y_2|z_2)],
\end{equation}
where $v_i\in V$ is an observation of input samples, and $z_i\in Z$ is an intermediate representation from the teacher network. $p(y_1|z_1)$ and $p(y_2|z_2)$ denote the predicted distributions on various samples.

\section{Distillation With Representative Learning Algorithms}\label{sec:architecture}
In this section, we will introduce distillation with multiple representative learning algorithms under the framework of Teacher-Student architectures.

\subsection{Multi-Teacher Distillation}

In the classic single teacher-based architecture, one student network is trained to match the ground truth as well as the transferred knowledge from one teacher network (e.g., soft logits and intermediate feature embeddings).
For example, MetaDistill~\cite{zhou2021meta} improves the knowledge transmission ability of a teacher network using meta-learning. Specifically, they introduce a pilot update mechanism to formulate the training of a teacher and a student as a bi-level optimization problem, so that the teacher can better transfer knowledge by exploiting feedback about the learning process of students.
SFTN~\cite{park2021learning} first modularizes a teacher and a student network into multiple blocks. The intermediate feature representations from each teacher block are followed by multiple student blocks, respectively, where the teacher and the student are simultaneously trained by minimizing the differences in the feature representations and logits between the teacher and the student. 
Xu et al.~\cite{xu2020knowledge} utilize self-supervised learning, when treated as an auxiliary task, to help gain more rounded and dark knowledge from a teacher network. Specifically, contrastive prediction is selected as the self-supervision task to maximize the agreement between a data point and its transformed version via a contrastive loss in latent space. 
Since the large gap between the teacher and the student could degrade the student performance, Mirzadeh et al.~\cite{mirzadeh2020improved} first introduce the teacher assistant (i.e, intermediate-sized network) as a multi-step teacher-student learning to eliminate this gap.
Bergmann et al.~\cite{bergmann2020uninformed} embrace multiple student networks that are simultaneously supervised with a power teacher network pre-trained on a large dataset of natural images, resulting in the accurate pixel-precise anomaly segmentation in high-resolution images. Thus, anomalies can be detected when the students fail to imitate the output of the teacher.

\begin{figure}[t]
\begin{subfigure}{0.45\textwidth}
  \centering
  \includegraphics[width=\textwidth]{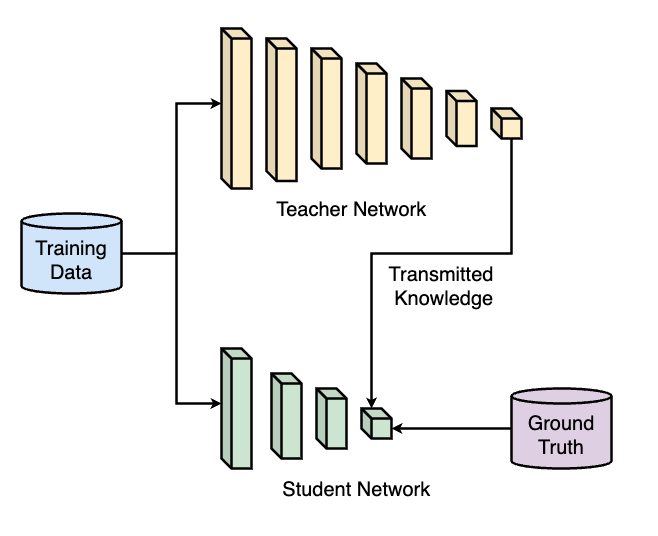}
  \caption{Knowledge learning from single teacher network.}
  \label{fig:single teacher}
\end{subfigure}
\begin{subfigure}{0.45\textwidth}
  \centering
  \includegraphics[width=0.95\textwidth]{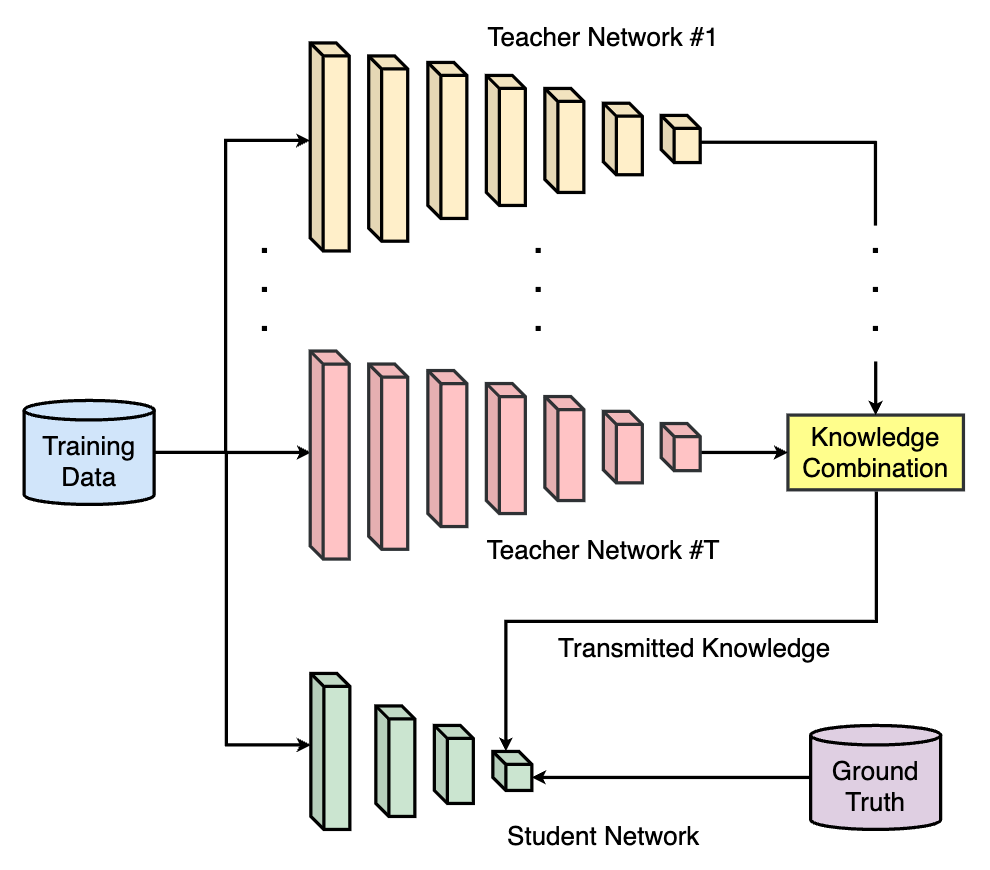}
  \caption{Knowledge learning from multiple teacher networks.}
  \label{fig:multiple teachers}
\end{subfigure}
\caption{The architecture of student network with the guided knowledge from single and multiple teacher networks.}
\label{fig:architecture}
\vspace{-5mm}
\end{figure}

Inspired by recent efforts, multiple teachers have been widely introduced in KD, where a student simultaneously receives knowledge transferred from multiple teachers. Consequently, a student network can robustly learn comprehensive and different knowledge under the guidance of multiple teacher networks. 
Fig.~\ref{fig:architecture} compares the architectures of student network with single and multiple teacher networks. In the multi-teacher distillation, averaging multiple teachers~\cite{tarvainen2017mean, furlanello2018born, 9287227, yang2020model, papernot2016semi} is a commonly-used approach to incorporate the potentially diverse knowledge from teachers (i.e., each teacher with an identical importance weight); concretely, a student network aims to learn the average softened logits of multiple teacher networks.

Note that multiple teacher networks can be heterogeneous since these teacher networks can be trained in various environments (e.g., different data distributions).  
This suggests that the transferred knowledge from various teachers can contribute differently to the student learning performance, so that the student network may learn more knowledge from similar teacher networks.
As a result, averaging multiple teacher networks could be sub-optimal by assigning each teacher an identical importance weight. 
Thus, to learn more representative and critical knowledge from significant teacher networks, advanced teacher weighting approaches have been introduced to assign a specific importance weight to each teacher network. 
For example, Adaptive Multi-teacher Multi-level Knowledge Distillation (AMTML-KD)~\cite{liu2020adaptive} includes multiple teacher networks, where each teacher network is learned an instance-level importance weight for adaptively integrating the intermediate feature representations from all teachers. Consequently, a student network can fully learn potentially diverse knowledge from multiple teachers. 
In addition to the logits from multiple teachers, You et al.~\cite{you2017learning} also additionally consider the relative similarity between intermediate representations of samples as one type of dark knowledge to guide the training of a student network. Concretely, the triplets are utilized to encourage the consistency of relative similarity relationships between the student and the teachers.
Yuan et al.~\cite{yuan2021reinforced} formulate the teacher selection problem under an RL framework, where each teacher network is assigned an appropriate importance weight based on various training samples and the outputs of teacher networks. Then, multiple teacher networks are randomly selected based on the learned importance weights to guide the training of the student network at each epoch. 
Ruder et al.~\cite{ruder2017knowledge} consider multi-teacher knowledge distillation for the domain adaptation and design teacher importance weights according to the data similarity between source domains and a target domain. Specifically, each teacher network is trained on a source domain, and the teacher importance weights are increased on the source domains similar to the target domain. 
\citet{Wu_2019_CVPR} simultaneously learn teacher importance weights by minimizing the validation empirical risk loss of the updated feature representations.

\subsection{Graph-Based Distillation}


With the recent advancements of Graph Neural Networks (GNNs) and geometric representations, graph-based distillation methods have been proposed to fully explore the structural relations of the data. 
To enhance the structural awareness of the student network, knowledge is represented as a graph~\cite{Liu_2019_CVPR, 9053986, Zhou_2021_ICCV, zhang2018better, Hou_2020_CVPR}, where vertex attributes denote the intermediate feature representations of training samples, and edge weights denote the similarity (e.g., Euclidean distance and cosine similarity) between the representations of two samples.
Specifically, \citet{Liu_2019_CVPR} transfer the intermediate vertex and edge transformations from the teacher to the student. \citet{9053986} and \citet{zhang2018better} learn the graph-based knowledge by minimizing the distance between the adjacency matrices of knowledge graphs formulated from the teacher and the student. 
\citet{Zhou_2021_ICCV} extract the geometric representations through Topology Adaptive Graph Convolutional Network (TAGCN)~\cite{du2017topology, kipf2016semi} on the attributed context graphs, and then maximize the mutual information (i.e., InfoNCE estimator~\cite{oord2018representation}) between such the graph-based knowledge. 
To transfer the scene structural knowledge in the road marking segmentation, \citet{Hou_2020_CVPR} introduce three inter-region affinity graphs (i.e., mean, variance, and skewness graph), in which a vertex attribute is the feature distribution statistic for each region computed by moment pooling~\cite{Peng_2019_ICCV, zellinger2017central}, and an edge weight denotes the cosine similarity between the feature distributions of two vertices. 


In addition, several studies have applied teacher-student architectures to GNNs, which build a shallower student GNN by distilling knowledge from a deeper teacher GNN~\cite{yang2021extract, yan2020tinygnn}. 
Given the feature embeddings of the node and the graph from teacher and student GNNs, \citet{Yang_2020_CVPR} introduce a Local Structure Preserving (LSP) module to explain the graphical semantics, which first generates the distribution for each local structure from both the teacher and student and then trains the student to learn the topological structure from the teacher by minimizing the distance between the distributions.
To preserve global structural information, \citet{9969463} introduce a node-level contrastive distillation task on pair-wise relations across the embedding spaces of the teacher and student, which trains the student to spatially align its node embeddings with the corresponding embeddings from the teacher (i.e., positive samples). 
\citet{deng2021graph} learn the fake topological graphs from the pre-trained teacher GNN by modeling the topology with a multivariate Bernoulli distribution, which is further transferred to the student GNN. 
\citet{feng2022freekd} collaboratively learn two shallower GNNs to distill the knowledge from each other in a hierarchical manner, which formulates the distillation for different nodes as a sequential decision-making problem. To solve this problem, this paper introduces the two-level distillation: (1) node-level distillation to distinguish which GNN distills knowledge for each node. (2) structure-level distillation to decide the local neighborhood subset of each node to be propagated. Once the direction and the local neighborhoods are determined, both the soft label and the neighborhood relations of each node are transferred to the other GNN.

\subsection{Federated Distillation}

Federated learning (FL) is a distributed machine learning approach that enables model training on many clients (for example, mobile devices). In this setup, the data is stored locally and does not need to be sent to a centralized location. Instead, models are sent to the data, trained locally, and the model updates (not the data) are sent back to a central server where they are aggregated to update the global model. It is especially beneficial when data privacy is crucial or impractical to send large amounts of data to a centralized location. While the parameter-averaging aggregation strategies in federated learning (FedAvg) effectively address data privacy concerns, they grapple with two key challenges: private data issues and heterogeneous models. The characteristics of private data may not align with those of other clients, leading to local models diverging from one another during training and fine-tuning on private instances. The clients might prefer not to reveal details of their models, rendering this parameter-sharing approach unsuitable.

Federated distillation methods have been effectively employed to address data heterogeneity issues, which can be applied either on the server side (e.g., utilizing a proxy dataset for ensemble distillation~\cite{lin2020ensemble_30, sattler2021fedaux_41, chen2020fedbe_7}) to adjust the global model, or on the client side by incorporating regularization techniques to control the data drift ~\cite{yao2021local_52,lee2021preservation_25,he2022learning_16,he2022class_17}. 
Specifically, \citet{chen2020fedbe_7} use a Bayesian model ensemble for robust aggregation as an alternative to average predictions. 
\citet{lin2020ensemble_30} present a server-side ensemble distillation method using a proxy dataset to support model heterogeneity and enhance FedAvg. 
The work of FedAUX~\cite{sattler2021fedaux_41} propose unsupervised pre-training on auxiliary data for client-side feature extractor initialization and ensemble prediction weighting based on private certainty scores. 
\citet{zhang2022fine_54} refine the global model server-side using data-free knowledge distillation and adversarial training of a generator model. The studies of \cite{lee2021preservation_25,yao2021local_52} have determined that local KD-based regularization effectively reduces the influence of non-IID data in FL scenarios. The client's local objective function is a mix of cross-entropy loss and a KD-based loss, which measures the discrepancy between the global (teacher) and local (student) models' outputs on private data, for example, through KL divergence.

For the heterogeneous models, the FedAvg protocol can be enhanced through server-side ensemble distillation during the aggregation process~\cite{lin2020ensemble_30,sattler2021fedaux_41}. The server manages prototypical models, aggregates post-update, and leverages unlabeled/synthetic data for fine-tuning, which enables effective knowledge exchange across diverse client architectures. 
The co-distillation adaptations serve as an alternative to FedAvg's parameter averaging. Several methods involve server aggregation of global targets updated by clients who distill locally and fine-tune models using their data. More modern strategies also consider server-side distillation \cite{sattler2021cfd_42,hu2021mhat_19,cheng2021fedgems_8}. The algorithm design is influenced by whether a labeled or unlabeled proxy dataset is chosen.
Different from the study~\cite{sattler2021cfd_42} compressing soft targets to boost communication, \citet{chang2019cronus_6} improve the efficiency by merging local distillation and training. The work of FedMD \cite{li2019fedmd_26} uses labeled datasets for pre-training, whereas ERA \cite{itahara2021distillation_20} adjusts aggregation to optimize training in non-IID settings. 

\subsection{Cross-Modal Distillation}
In cross-modal learning environments, there might be scenarios where certain modalities' data or labels are unavailable during training or testing \citep{DBLP:conf/cvpr/GuptaHM16, DBLP:conf/eccv/GarciaMM18}. 
This poses a challenge and accentuates the necessity for knowledge transfer between different modalities.
The integration of KD into such environments represents a significant stride, facilitating an enhanced understanding and synergy of varied data modalities.

\citet{bucilua2006model} and \citet{ba2014deep} make earlier contributions to this field, where they explore the principles of model compression and proposed effective ways to distill the knowledge of a large model into a smaller one.
\citet{luo2019knowledge} later explore the interplay between KD and cross-modal learning, introducing a novel technique that capitalizes on the unique strengths of distinct modalities in a KD setting, specifically addressing the text and visual modalities. 
Their method integrates the essence of KD, in tandem with the intricacies of cross-modal data handling, thereby paving the way for future advancements in the field. 
Subsequently, drawing on the groundwork of \citet{luo2019knowledge}, \citet{kim2020deep} propose a unique approach known as deep cross-modal projection learning for image-text embeddings. 
Their innovative KD method involves transferring knowledge from visual to textual representation, and vice versa, showcasing its proficiency in text-to-image synthesis tasks. 
This work significantly expands the purview of KD in the realm of cross-modal learning. 
In the same year, \citet{tian2020improving} demonstrate how KD could enhance cross-modal sentiment analysis performance. 
Their research marks a notable instance of applying KD to cross-modal sentiment analysis, an area necessitating the understanding and synthesis of information from different modalities, which sets a precedent for future studies, especially those focusing on emotion-based cross-modal analysis. 

The potential of KD in cross-modal learning is further explored by \citet{nguyen2021multimodal}, who focus on video and language processing. 
Their research demonstrates how KD, when aptly incorporated, could significantly improve the performance of models working on these complex, cross-modal tasks. 
The implications of their work extend to applications such as video summarization, captioning, and more. 
Other noteworthy works like \cite{guo2019mixture}, \cite{huang2020leveraging}, \cite{gao2020multimodal}, \cite{wang2020survey}, \cite{zoph2021self}, and \cite{zhou2021large} have significantly advanced the state of cross-modal distillation, showing its effectiveness in various fields ranging from machine translation to surveillance systems, and from medical imaging to dialog systems. 

\section{Distillation Schemes}\label{sec:learning scheme}
In this section, we will summarize online distillation and self-distillation schemes under the framework of Teacher-Student architectures, which suggest that both student and teacher networks will be learned during the training stage.

\subsection{Online Distillation}

Table~\ref{tab:learning scheme} compares the different distillation schemes in terms of teacher and student learning statuses, as well as their role statuses. 
The classic distillation scheme is offline distillation~\cite{hinton2015distilling}, which represents that student networks learn the knowledge transferred from pre-trained teacher networks. Specifically, powerful teacher networks are first completely well-trained on large-scale datasets, and then transfer the knowledge to guide the training of compact student networks. The roles of teacher and student networks are not exchanged during the student training process. 
It is important to note that offline distillation is not included in the scope of this survey, and we refer readers to the existing surveys that have provided comprehensive reviews on the offline distillation~\cite{gouKnowledgeDistillationSurvey2021, alkhulaifi2021knowledge}. 

Although the offline distillation scheme is straightforward and computationally efficient, powerful pre-trained teacher networks and large-scale datasets could be limited~\cite{mirzadeh2020improved, zhang2018deep, chen2020online}. To solve such issues, an online distillation scheme is further commonly utilized to simultaneously train student and teacher networks, so that the whole knowledge learning process can be end-to-end trainable.
For instance, in On-the-fly Native Ensemble (ONE) learning strategy~\cite{zhu2018knowledge, li2021online}, a native ensemble teacher network is created from multiple student branches on-the-fly and simultaneously trained with these student branches. With student branches, both teachers and students are more efficient to be trained with superior generalization performance and without asynchronous model updates.
Due to the gating module on the shared layer, student branches are limited to the same network architecture in ONE learning strategy. To overcome this limitation, Feature Fusion Learning (FFL)~\cite{kim2021feature} is further proposed to allow student branches to be applicable to any architecture.
To better boost the knowledge distillation process, Su et al.~\cite{su2021attention} additionally introduce an attention mechanism to capture important and high-level knowledge, so that teachers and students can be dynamically and effectively trained with the help of the valuable knowledge.  
In Peer Collaborative Learning (PCL)~\cite{wu2021peer}, a peer ensemble teacher is trained with the ensemble feature representation of multiple  student peers. Besides, a temporal peer means the teacher is individually built for each student peer to collaboratively transfer knowledge among student peers. 

\begin{table}[t]
\renewcommand{\arraystretch}{1.10}
\newcommand{\tabincell}[2]{\begin{tabular}{@{}#1@{}}#2\end{tabular}}
\centering
\caption{The comparison between the different distillation schemes.}\label{tab:learning scheme}
\begin{tabular}{c|c|c|c}
\toprule
\tabincell{c}{Distillation \\ scheme} & \tabincell{c}{Teacher \\ training \\ status} & \tabincell{c}{Student \\ training \\ status} & \tabincell{c}{Role \\ status} \\
\midrule
\tabincell{c}{Offline distillation \\ \cite{hinton2015distilling, romero2014fitnets, tang2019distilling, yim2017gift, wang2018kdgan} \\ \cite{li2017large, tang2018ranking, zhang2018deep, furlanello2018born, Zhu_2021_CVPR} \\ \cite{tian2019contrastive, Shrivastava_2023_CVPR, Passalis_2018_ECCV, Ahn_2019_CVPR, sohn2020simple, wang2021data} \\ \cite{li2019bidirectional, tsai2018learning, hoffman2018cycada, ghiasi2021multi, yang2022cross, li2020knowledge}}  & \tabincell{c}{Well trained} & \tabincell{c}{To be trained} & Static \\
\hline
\tabincell{c}{Online distillation \\ \cite{mirzadeh2020improved, zhang2018deep, chen2020online, zhu2018knowledge, li2021online} \\ \cite{kim2021feature, Passalis_2020_CVPR, su2021attention, wu2021peer, matiisen2019teacher} \\ \cite{Isobe_2021_CVPR, 9206989, Xue_2021_ICCV, Chelaramani_2021_WACV}} & \tabincell{c}{To be trained} & \tabincell{c}{To be trained} & Static \\ 
\hline
\tabincell{c}{Self-distillation \\ \cite{yuan2020revisiting, pang2020self, chung2020feature, chen2020online, xie2020self} \\ \cite{yang2021multi, tzelepi2021online, guo2020online, Tian_2021_CVPR}}  & \tabincell{c}{To be trained} & \tabincell{c}{To be trained} & Dynamic \\
\hline
\end{tabular}
\end{table}

\subsection{Self-Distillation}

The self-distillation scheme is one particular scheme of online distillation. Different from the classic online distillation methods, the roles of student and teacher networks are dynamic during the iterative training process, which indicates that student and teacher networks can be exchanged or student networks are able to learn the knowledge from themselves (i.e., Teacher-free knowledge learning) in the self-distillation scheme. 
Yuan et al.~\cite{yuan2020revisiting} demonstrate KD as a type of label smoothing regularization, and thus label smoothing regularization can be further regarded as a virtual teacher. Based on these analyses, Teacher-free Knowledge Distillation is further proposed, where a student can learn the knowledge from itself or a manually-designed regularization distribution. 
Pang et al.~\cite{pang2020self} further leverage the iterative self-distillation scheme for the end-to-end video anomaly detection, where the student networks (i.e., the anomaly learners) continuously generate the new pseudo labels to replace the previous ones. 
Instead of a powerful teacher network, OKDDip consists of multiple auxiliary student peers and one group leader in the two-level distillation~\cite{chen2020online}. Each student peer is self-trained on the knowledge aggregated from other peers with different importance weights. Then all peers further guide the training of the group leader during the second distillation. 
Chung et al.~\cite{chung2020feature} transfer the feature maps and the soft logits through the online adversarial knowledge distillation, where each network has a discriminator that distinguishes the feature maps from its own and other networks. Through the adversarial training of the discriminator, each network can learn the distribution of feature maps from other networks by fooling the corresponding discriminator. 
Multi-view Contrastive Learning (MCL)~\cite{yang2021multi} captures the correlation of feature representations of multiple student peers, and each peer provides a unique feature representation that suggests a specific view of input data. These feature representations can be regarded as more powerful knowledge to be distilled for the effective training of student peers.
Online Subclass Knowledge Distillation (OSKD)~\cite{tzelepi2021online} reveals the similarities inside each class to capture the shared semantic information among subclasses. During the online distillation process, each sample moves closer to the representations of the same subclass while further away from that of different subclasses. 
Recent online teacher-student learning benefits from collaborative learning, where mutual knowledge is shared to enhance the learning ability of both teachers and students. 
In Knowledge Distillation via Collaborative Learning (KDCL)~\cite{guo2020online}, multiple students with different capacities are assembled to simultaneously provide high-quality soft logits to instruct themselves with significant performance improvement. Specifically, each student is fed with individually-distorted images to reduce variance against perturbations in the input data domain. 
\section{Applications}\label{sec:application}

The applications of Teacher-Student architectures with KD are introduced in this section. Specifically, we categorize the applications based on various purposes of networks: classification, recognition, generation, ranking, and regression. 


\subsection{Classification Purpose}

\subsubsection{CV}
Image classification issues can be resolved through KD~\cite{zhuLowresolutionVisualRecognition2019,bagherinezhadLabelRefineryImproving2018,pengCorrelationCongruenceKnowledge2019,liLearningForgetting2017}. A label refinement approach for self-improvement and label augmentation is proposed \cite{bagherinezhadLabelRefineryImproving2018} for incomplete, ambiguous, and redundant image labels in order to learn soft, informative, collective, and dynamic for complicated image classification labels. In order to classify low-resolution images, Zhu et al. \cite{zhuLowresolutionVisualRecognition2019} present deep feature distillation, in which the student's output features are compared to the teacher's output features. This method was inspired by KD-based low-resolution face recognition.

\subsubsection{NLP}
Text classification tasks include sentence categorization and sentence classification. Included are broad domain knowledge and knowledge distillation relevant to a certain task. Tang et al. \cite{tangDistillingTaskspecificKnowledge2019} suggest task-specific KD from a BERT teacher model to a bidirectional long short-term memory network for sentence classification and matching. In order to learn across different NLP tasks, a lightweight student model called DistilBERT \cite{sanhDistilbertDistilledVersion2019} that has the same basic structure as BERT was created. The authors of \cite{aguilarKnowledgeDistillationInternal2020} suggest internal distillation to create a condensed version of the huge teacher BERT called the student BERT.

\subsubsection{Speech}
For audio classification, to maximize information transmission, \cite{gaoAdversarialFeatureDistillation2019} demonstrates a multi-stage feature distillation process and used an adversarial learning strategy. KD is employed as a voice improvement technique to increase noise-robust speech recognition, and an audio-visual multi-modal KD approach is developed \cite{perezAudiovisualModelDistillation2020}. The student model for audio data transfers knowledge from the teacher model for visual and auditory data. In essence, this improvement allows teachers and students to communicate cross-modal information \cite{perezAudiovisualModelDistillation2020,albanieEmotionRecognitionSpeech2018,rohedaCrossmodalityDistillationCase2018}. Shi et al. \cite{shiCompressionAcousticEvent2019} suggest a quantitative distillation approach that combines KD and quantification to effectively detect acoustic occurrences.

\subsubsection{Data Security and Privacy}

Adversarial samples can fool the behaviors of neural networks (e.g., misclassification) by maliciously manipulating the input data to the networks.
\citet{gil2019white} introduce a black-box adversarial attack that targets the toxicity classifier detecting toxic language on social media. The source classifier (i.e., the teacher) is first trained on the data with a similar distribution to the target data, and then adversarial samples are generated through HotFlip~\cite{ebrahimi2017hotflip} on the source classifier. The attacker (i.e., the student) is trained on such generated samples and performs a black-box attack against the target model.
To tackle the data security issue, several studies have introduced KD to build robust student networks against malicious perturbations. \citet{goldblum2020adversarially} encourage the student to mimic the teacher's predictions within an $\epsilon$-ball of training samples, by minimizing the discrepancy between the predictions of the teacher on normal images and the predictions of the student on adversarial images.
The prediction probability vector of the teacher indicates the explicit relative information about classes, which can improve the network generalization capability on unseen data~\cite{7546524}. Hence, to enhance the network resilience to perturbations, \citet{7546524} train the student solely on the soft probability vectors predicted by the teacher. 

Besides, the data privacy issue also poses significant challenges in providing robust and generalized networks with limited training data. 
To solve this challenge, \citet{bai2020few} adopt cross distillation to reduce the layer-wisely accumulated errors on the student network. In addition to receiving the layer-wise feature maps from the teacher to the student (i.e., the correlation), the student's intermediate feature maps are also transferred to the teacher (i.e., the imitation). To balance the correlation and imitation, \citet{bai2020few} further introduce the convex combination between their loss functions or the intermediate representations of the teacher and student networks. 
Instead of original training data, \citet{lopes2017data} train the student network on the activation records~\cite{Mahendran_2015_CVPR} and reconstructed data from the teacher network. Specifically, to reconstruct original training data, this work~\cite{lopes2017data} inputs the random Gaussian noise into the teacher, and then applies gradients to iteratively minimize the difference between the activation records and those for the noisy image.
\citet{papernot2016semi} introduce the privacy protection strategy for training data by limiting the student training to the topmost teacher votes after adding random noise. This strategy involves multi-teacher training on disjoint subsets of sensitive data, and analyzes the sensitivity of each teacher vote through the moment accountant technique~\cite{abadi2016deep}. With the auxiliary and non-sensitive data, the student network is trained on the partially-aggregated predictions of the teachers with the topmost votes.

\subsection{Recognition Purpose}

\subsubsection{CV}
One of the applications of knowledge learning is to enhance face recognition and improve the performance of the model from the perspective of efficiency and accuracy.  For example, in \cite{luoFaceModelCompression2016}, the student network of chosen information neurons at the top layer of the teacher network receives knowledge from the teacher network. The teacher-weighted technique for knowledge transfer eliminated the feature representation of the hint layer in order to prevent the teacher from accidentally leading students astray \cite{wangAdversarialLearningPortable2018}. A technique of recursive KD is developed that initializes the newer networks using the previous student network \cite{yanVargfacenetEfficientVariable2019}. Since the majority of face recognition techniques use a test set for which the training set is unaware of the category or identity, such as \cite{duongShrinkTeaNetMillionscaleLightweight2019} describes the angle loss, the distance between positive and negative sample features is often the face recognition criteria.

\subsubsection{NLP}
Natural language processing has been substantially improved by BERT  \cite{devlinBertPretrainingDeep2019}, an expansive and complex multilingual model that is also challenging to train. To address the issue of the training which takes a lot of time and resources, \cite{sunPatientKnowledgeDistillation2019,jiaoTinybertDistillingBert2020,tangDistillingTaskspecificKnowledge2019,wangImprovedKnowledgeDistillation2019} propose various compact versions of BERT utilizing knowledge distillation. Jiao et al. \cite{jiaoTinybertDistillingBert2020} propose TinyBERT, a transformer-based KD, to accelerate the inference speed. For sentiment analysis, paraphrase similarity matching, natural language inference, and machine reading comprehension, Sun et al. \cite{sunPatientKnowledgeDistillation2019} suggest KD, in which the teacher hint layer's feature representation of the labels is transmitted to the students in this strategy.

\subsubsection{Speech}
The certain challenging issues in speech recognition can be resolved through several KD studies~\cite{baiLearnSpellingTeachers2019,asamiDomainAdaptationDNN2017,ghorbaniAdvancingMultiaccentedLstmctc2018}. 
To address the issue of acoustic model overfitting in sparse data, \citet{asamiDomainAdaptationDNN2017} employ KD as a regularisation technique to train an adaptive model under the direction of the source model. 
\citet{ghorbaniAdvancingMultiaccentedLstmctc2018} train an advanced multi-accent student model by transferring knowledge across domains and numerous accent-specific RNNs.

\subsection{Generation Purpose}

\subsubsection{CV}

A feature map-based KD approach~\cite{chenAdversarialDistillationEfficient2019} for GANs is proposed to increase the accuracy of image categorization. Students receive knowledge from feature maps in this way. By combining a deep generative model for diagnosis with a teacher-student model for interpretation, \citet{wangAdversarialLearningPortable2018} design a visual interpretation and diagnosis framework for an image classifier.

\subsubsection{NLP}
One of KD-based applications in the generative model of natural language processing is neural machine translation. However, high-performing NMT models are frequently enormous and complex, like BERT models. To generate lightweight NMT models, many extended KD techniques have been proposed~\cite{hahnSelfknow2019, zhouM2KDMultimodelMultilevel2020, kimSequencelevelKnowledgeDistillation2016}. 
\citet{zhouM2KDMultimodelMultilevel2020} demonstrate how the capabilities of KD-based NAT models and the data acquired through knowledge transfer are crucial to their superior performance. 
\citet{gordonExplainingSequencelevelKnowledge2019} explain the positive aspects of sequence-level KD from the standpoint of data augmentation and regularisation. 

Effective word-level KD~\cite{kim2021feature} is extended to sequence level for the sequence creation scenario of NMT. The distribution of sequences produced by the student model is similar to that of the instructor. \citet{tanEfficientNetRethinkingModel2019} suggest multi-teacher distillation to handle the issue of linguistic diversity, where numerous individual models dealing with bilingual couples are instructors and multilingual models are pupils.
\citet{weiOnlineDistillingCheckpoints2019} suggest that the model that performs the best during training is picked as the teacher and updated by any following better model in order to enhance the performance of machine translation tasks.

\subsection{Ranking Purpose}
In the domain of recommender systems, the crucial task is predominantly a ranking problem with the objective of accurately and efficiently predicting user-preferred items in an ordered sequence~\cite{LTR_RecSys}. 
To fulfill this task, KD, a technique that transfers knowledge from a complex teacher model to a compact student model, is harnessed, enabling a reduction in computational demands while preserving or enhancing performance. 

Reviewing KD literature reveals diverse approaches to addressing ranking-related challenges in the recommendation. 
For instance, \citet{zhou2018rocket} propose a rocket launching process, which simultaneously trains the light network (i.e., the student) and the booster network (the teacher) by learning the predictions from each other. Note that some lower-level layers are shared between these two networks. 
\citet{zhu2020ensembled} introduce an adaptive ensemble distillation framework for click-through prediction, where the importance weights of teacher networks are learned through a teacher gating network in a sample-wise manner. To avoid overfitting when training the student network, the distillation loss from the teacher network is further considered as the early-stopping criteria. 
\citet{KD_rec_topn} propose a novel KD technique for collaborative filtering, which incorporates rank-aware sampling and unique training strategies to alleviate the effects of data sparsity and ambiguous feedback. 
To extend this, \citet{DE-RRD} suggest the student model's learning should not only mirror the teacher's predictions but also integrate the latent knowledge embedded in the teacher model. Their approach utilizes Distillation Experts and a Relaxed Ranking Distillation method to transfer knowledge, considering items' relaxed ranking orders.
Moreover, \citet{GeneralKD_counterfactual} introduce a counterfactual recommendation methodology employing various distillation strategies including label-based, feature-based, sample-based, and model structure-based to manage inherent biases and optimize performance in recommender systems. 
Similarly, \citet{Topology_KD} advocate for transferring the topological structure rather than solely point-wise representation from the teacher model to the student, enabling the capture of the relational information integral to recommender systems. 
Finally, \citet{UnbiasKD} put forward a stratified distillation strategy aiming to correct popularity biases and enhance the equity of recommendations.

\subsection{Regression Purpose}

To address regression problems, several approaches have been proposed in the literature. 
\citet{9175560} introduce a novel teacher outlier rejection loss function that identifies outliers in training samples using teacher predictions. They employ a multi-task network with two outputs: the ground truth estimation and the teacher prediction, effectively training a student network. 
Another method for regression tasks is the data-free KD technique proposed by \citet{DBLP:journals/eswa/KangK21}. 
This method utilizes a generator network to generate synthetic data points, enabling the student network to replicate the prediction ability of the teacher network without requiring a training dataset. 
Experimental results demonstrate the superior prediction accuracy and lower RMSE achieved by this approach compared to baseline methods on various benchmark datasets.

In the context of representation learning, \citet{DBLP:conf/iclr/0038MBT21} leverage a Softmax Regression Loss to optimize the student network's output feature, emphasizing effective representation learning. 
They introduce two loss functions: the first focuses on direct feature matching to optimize the student's penultimate layer feature, while the second decouples representation learning and classification by utilizing the teacher's pre-trained classifier. 
These loss functions effectively transfer relevant information while preserving the representational power of the student's feature.

Furthermore, for time-series regression tasks in industrial manufacturing, \citet{TimeSeries_Regression} propose Contrastive Adversarial Knowledge Distillation (CAKD). 
This approach employs adversarial and contrastive learning to transfer knowledge from a complex Long Short-Term Memory (LSTM) model to a simpler Convolutional Neural Network (CNN) model. 
The CAKD method enhances model performance and efficiency, making it more applicable to resource-limited environments.
These advancements in KD for regression demonstrate the potential of leveraging teacher networks and synthetic data to improve the performance, efficiency, and generalization capabilities of student networks in regression tasks.

\section{Opportunities and Future Works}\label{sec:future work}

In this section, we discuss the opportunities and potential directions for improving the distillation process from the following perspectives: the architecture design of teacher and student networks, the quality of knowledge, and the theoretical studies of regression-based knowledge learning.

\subsection{Teacher-Student Architecture Design}

To improve the distillation process, some studies have investigated the relationship between the Teacher-Student architecture design and learning performance. 
The knowledge is commonly represented using soft labels and feature embeddings from teachers, and bigger and more robust teacher networks can be expected to provide more representative and reliable knowledge. Therefore, the classic approach to train more accurate students is to design large-scale teacher networks. 
However, some recent works investigate that the increasing gap (in size) between student and teacher networks could decrease the learning performance~\cite{mirzadeh2020improved, zhang2019your, kang2020towards}. 
Recently, Neural Architecture Search (NAS) has witnessed many successes in automatically designing neural networks in solving some tasks, such as image classification~\cite{liu2018progressive}, NLP~\cite{wang2020textnas}, speech recognition~\cite{mehrotra2020bench}, among others. 
For example, Kang et al.~\cite{kang2020towards} integrate NAS with the oracle KD to find the optimal student network structures and operations for potentially achieving better performance than teacher networks. 
With the help of NAS, the efficient student network can be found from the fixed and high-performing teacher network with lower computational cost and fewer network parameters~\cite{bashivan2019teacher}.  
Hence, NAS can be further incorporated with KD in the future direction, which searches for the best pair of powerful teacher and combat student networks, leading to efficient learning performance.

\subsection{Knowledge Quality}

Existing works have been commonly dedicated to improving the distillation process by designing efficient network structures and transferring better feature representations. 
However, there are fewer studies analyzing the amount of informative knowledge that can be potentially utilized and transferred from teachers. 
Through the quantification of visual concepts encoded in teacher networks, Cheng et al.~\cite{cheng2020explaining} explain the success of KD from the three hypotheses: learning more visual concepts, learning various visual concepts, and yielding more stable optimization directions. 
Miles et al.~\cite{miles2021information} integrate the analysis of information theory with KD using infinitely divisible kernels, which achieve the computationally-efficient distillation process on the cross-model transfer tasks. 
Therefore, the quantification of knowledge can be investigated in the future research direction, which aims to analyze how much important knowledge can be potentially captured before the learning process.

\subsection{Theoretical Understandings of Regression-Based Knowledge Learning}

Currently, most teacher-student architectures are employed on classification tasks, where intermediate feature embeddings and soft logits can be commonly represented as dark knowledge transferred to student networks. Moreover, more advanced training schemes and architecture designs are introduced to improve the efficiency of the distillation process. 
Although several works\cite{9175560, DBLP:journals/eswa/KangK21, DBLP:conf/iclr/0038MBT21, TimeSeries_Regression} focus on regression tasks, one promising research direction can be investigated in the theoretical studies of regression-based knowledge learning, such as the representation of dark knowledge on regression problems.
The final predictions of teacher networks are represented as knowledge to be transferred to student networks~\cite{takamoto2020efficient, kang2021data}, and student networks also aim to mimic the extracted feature representations from teacher networks~\cite{xu2022contrastive, saputra2019distilling}. 
With deeper theoretical studies on regression-based knowledge learning, teacher-student architectures will be further effectively employed in practical applications.

\section{Conclusion}\label{sec:conclusion}

Teacher-Student architectures were first proposed in KD, which aim to obtain lightweight student networks with comparable performance to deep teacher networks.  Different from the existing KD surveys~\cite{gouKnowledgeDistillationSurvey2021, wang2021knowledge, alkhulaifi2021knowledge} primarily focusing on the objective of knowledge compression, this survey provides a comprehensive review of Teacher-Student architectures for multiple distillation objectives, including knowledge compression, knowledge expansion, knowledge adaptation, and knowledge enhancement. 
Moreover, this survey not only introduces various knowledge representations and optimization objectives associated with each specific representation, but also provides a detailed overview of representative learning algorithms and distillation schemes under the framework of Teacher-Student architectures. 
The recent applications of Teacher-Student architectures with KD are summarized based on various network purposes: classification, recognition, generation, ranking, and regression. 
Lastly, this survey investigates the promising research directions of knowledge learning on Teacher-Student architecture design, knowledge quality, and theoretical studies of regression-based learning, respectively. 
Through this comprehensive survey, industry practitioners and the academic community can gain valuable insights and guidelines for effectively designing, learning, and applying Teacher-Student architectures on various distillation objectives.

\bibliographystyle{IEEEtranN}
\bibliography{references}


\vspace{-20mm}
\begin{IEEEbiography}[{\includegraphics[width=0.85\columnwidth]{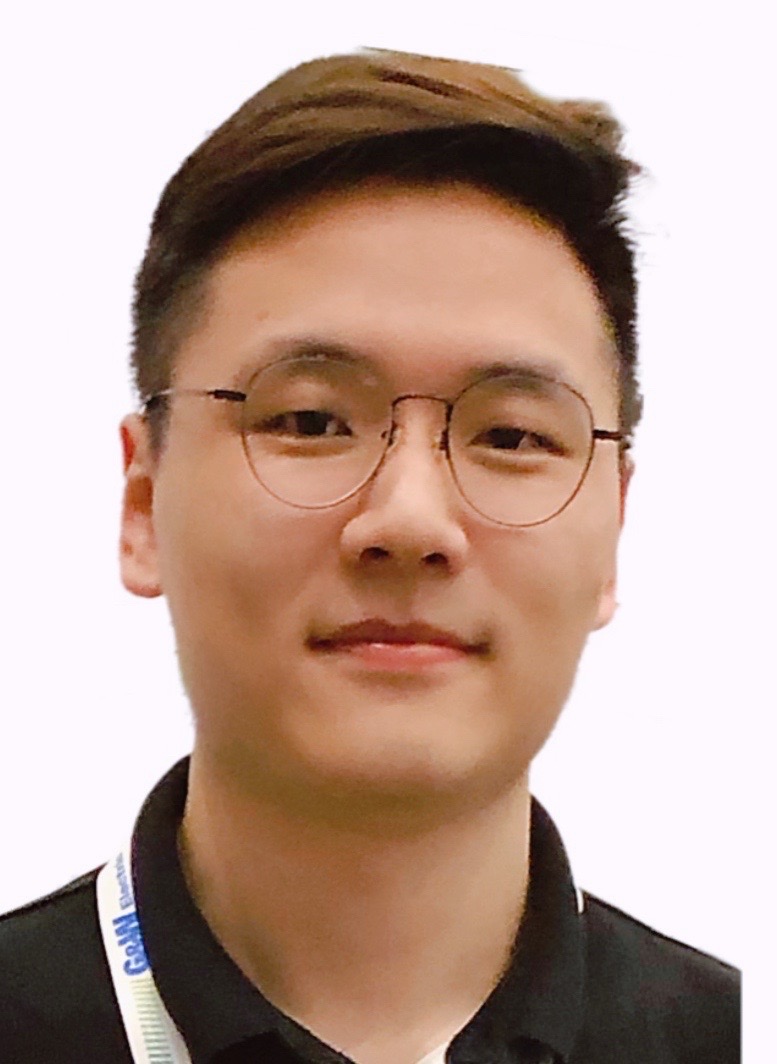}}]{Chengming Hu} is currently a Ph.D. candidate at McGill University, Canada. He received M.Sc. in Quality Systems Engineering with Concordia Institute for Information Systems Engineering (CIISE), Concordia University, Canada, in 2019. 

His research interests include computational intelligence and cyber-physical security, with applications in smart grids and wireless communication systems.  
\end{IEEEbiography}
\vspace{-20mm}

\begin{IEEEbiography}[{\includegraphics[width=0.85\columnwidth]{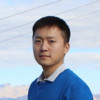}}]{Xuan Li} is currently a Ph.D. candidate at McGill University, Canada. His primary research focuses on Out-of-Distribution detection in medical imaging, unsupervised semantic segmentation, and Bayesian uncertainty learning. 

He also works closely with industry for general AI applications, including automatic airport apron management system, clinical skin lesion identification platform, etc.
\end{IEEEbiography}
\vspace{-15mm}

\begin{IEEEbiography}
[{\includegraphics[width=0.85\columnwidth]{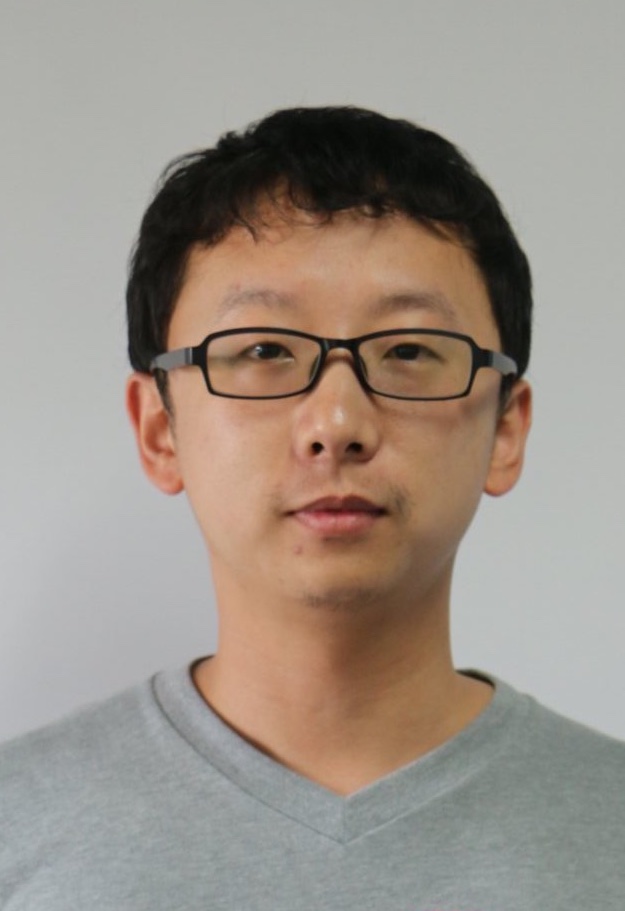}}]{Dan Liu} is a Ph.D. candidate at McGill University, Canada. His research focus includes model quantization, pruning and knowledge distillation.
\end{IEEEbiography}
\vspace{-10mm}

\begin{IEEEbiography}[{\includegraphics[width=0.85\columnwidth]{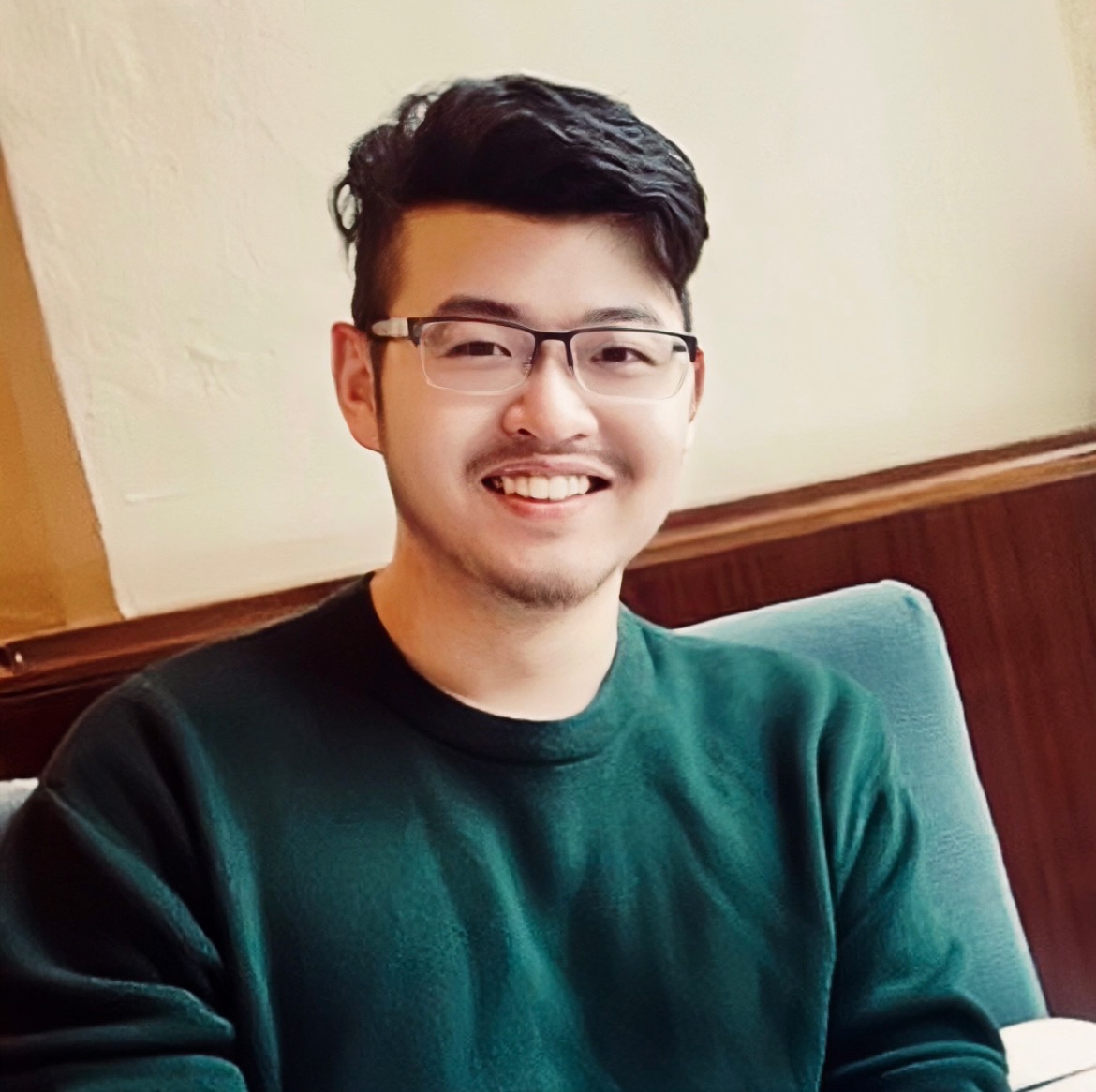}}]{Haolun Wu} is a Ph.D. candidate in Computer Science at Mila and McGill University.
He is broadly interested in knowledge modeling, retrieval, trustworthy ML, etc.
He is a recipient of Borealis AI Fellowship 2022-2023.
He has published papers on SIGIR, AAAI, CIKM, TOIS, ICDE, etc.
He interned / collaborated with Google Research,  Microsoft Research, Bell Canada, and Huawei Noah's Ark Lab.
He was invited to give talks at Mila and Microsoft Bing.
He served as a session chair at SIGIR 2022.
\end{IEEEbiography}
\vspace{-10mm}

\begin{IEEEbiography}
[{\includegraphics[width=0.85\columnwidth]{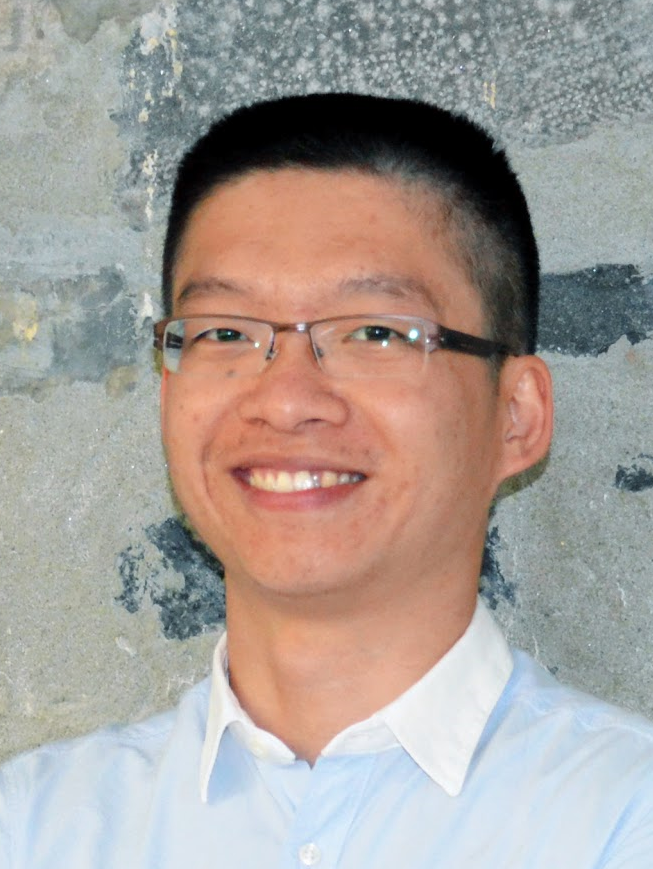}}]{Xi Chen} is currently a Senior Principle Researcher at Huawei Noah's Ark Lab, Montreal. He also serves as an adjunct professor at School of Computer Science, McGill University. His experience and passion lie in a wide range of AI domains, including AI for 5G/6G, AI for networking, prediction, smart IoT, WiFi sensing, NLP, self-driving, smart homes, smart systems, vehicle-to-everything, etc. He achieved his PhD degree at School of Computer Science, McGill University. He received both M.Eng. and B.S. degrees from Department of Electronic Engineering, Shanghai Jiao Tong University.
\end{IEEEbiography}
\vspace{-10mm}

\begin{IEEEbiography}[{\includegraphics[width=0.85\columnwidth]{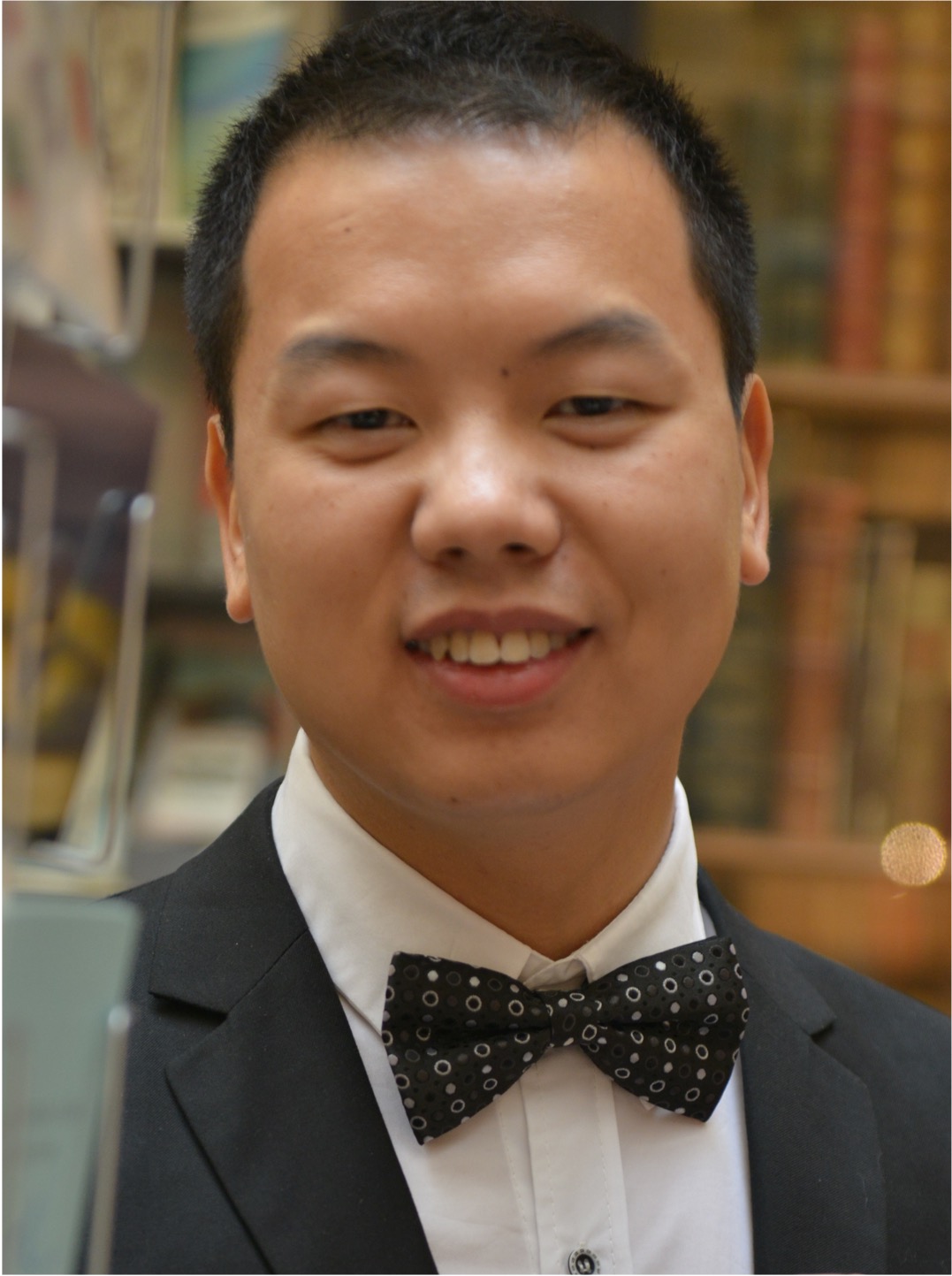}}]{Ju Wang} is a senior researcher at Samsung AI Center (SAIC)-Montreal since 2020. He also serves as an adjunct professor at Northwest University.
He did two-year postdoc at University of Waterloo during 2018-2020. He received his Ph.D. in Computer Science from Northwest University, China, in 2017.

His research interests include: Artificial Intelligence, Wireless Communication System/Network, and Internet of Things (IoT).
\end{IEEEbiography}

\begin{IEEEbiography}[{\includegraphics[width=0.9\columnwidth]{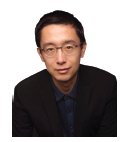}}]{Xue Liu (Fellow, IEEE)} received the B.S. degree in mathematics and the M.S. degree in automatic control from Tsinghua University, Beijing, China, in 1996 and 1999, respectively, and the Ph.D. degree in computer science from the University of Illinois at Urbana–Champaign, Champaign, IL, USA, in 2006. 

He is currently a Professor and a William Dawson Scholar with the School of Computer Science, McGill University, Montreal, QC, Canada. He was also the Samuel R. Thompson Associate Professor with the University of Nebraska-Lincoln and HP Labs, Palo Alto, CA, USA. He has been granted one U.S. patent and filed four other U.S. patents, and published more than 150 research papers in major peer-reviewed international journals and conference proceedings. His research interests include computer networks and communications, smart grid, real-time and embedded systems, cyber–physical systems, data centers, and software reliability. Dr. Liu was a recipient of several awards, including the Year 2008 Best Paper Award from the IEEE TRANSACTIONS ON INDUSTRIAL INFORMATICS, and the First Place Best Paper Award of the ACM Conference on Wireless Network Security in 2011. He received the Outstanding Young Canadian Computer Science Researcher Prizes from the Canadian Association of Computer Science.
\end{IEEEbiography}

\end{document}